# The Linearization of Belief Propagation on Pairwise Markov Random Fields

**Wolfgang Gatterbauer**
Carnegie Mellon University
Pittsburgh, Pennsylvania 15213

## Abstract

Belief Propagation (BP) is a widely used approximation for exact probabilistic inference in graphical models, such as Markov Random Fields (MRFs). In graphs with cycles, however, no exact convergence guarantees for BP are known, in general. For the case when all edges in the MRF carry the same symmetric, doubly stochastic potential, recent works have proposed to approximate BP by linearizing the update equations around default values, which was shown to work well for the problem of node classification. The present paper generalizes all prior work and derives an approach that approximates loopy BP on any pairwise MRF with the problem of solving a linear equation system. This approach combines exact convergence guarantees and a fast matrix implementation with the ability to *model heterogenous networks*. Experiments on synthetic graphs with planted edge potentials show that the linearization has comparable labeling accuracy as BP for graphs with weak potentials, while speeding-up inference by orders of magnitude.

## 1 Introduction

Belief Propagation (BP) is an iterative message-passing algorithm for performing inference in graphical models (GMs), such as Markov Random Fields (MRFs). BP calculates the marginal distribution for each unobserved node, conditional on any observed nodes (Pearl 1988). It achieves this by propagating the information from a few observed nodes throughout the network by iteratively passing information between neighboring nodes. It is known that when the graphical model has a tree structure, then BP converges to the true marginals (according to exact probabilistic inference) after a finite number of iterations. In loopy graphs, convergence to the correct marginals is not guaranteed; in fact, it is not guaranteed at all, and using BP can lead to well-documented convergence problems (Sen et al. 2008). While there is a lot of research on convergence of BP (Elidan, McGraw, and Koller 2006; Ihler, Fisher III, and Willsky 2005; Mooij and Kappen 2007), exact criteria for convergence are not known (Murphy 2012), and most existing bounds for BP on general pairwise MRFs give only sufficient convergence criteria, or are for restricted cases, such as when the underlying distributions are Gaussians (Malioutov, Johnson, and Willsky 2006; Su and Wu 2015; Weiss and Freeman 2001).

**Semi-supervised node classification.** BP is also a versatile formalism for semi-supervised learning; i.e., assigning classes to unlabeled nodes while maximizing the number of correctly labeled nodes (Koller and Friedman 2009, ch. 4). The goal is to predict the most probable class for each node in a network *independently*, which corresponds to the Maximum Marginal (MM) assignment (Domke 2013; Weiss 2000). Let $\mathbb{P}$ be a probability distribution over a set of random variables $\mathbf{X} \cup \mathbf{Y}$. MM-inference (or "MM decoding") searches for the most probable assignment $y_i$ for each unlabeled node $Y_i$ *independently*, given evidence $\mathbf{X} = \mathbf{x}$:

$$\mathrm{MM}(\mathbf{y}|\mathbf{x}) = \{\arg\max_{y_i} \mathbb{P}(Y_i = y_i | \mathbf{X} = \mathbf{x}) | Y_i \in \mathbf{Y}\}$$

Notice that this problem is simpler than finding the actual marginal distribution. It is also *different* from finding the Maximum A-Posteriori (MAP) assignment (the "most probable configuration"), which is the mode or the most probable *joint classification* of all non-evidence variables:[1]

$$\mathrm{MAP}(\mathbf{y}|\mathbf{x}) = \arg\max_{\mathbf{y}} \mathbb{P}(\mathbf{Y} = \mathbf{y} | \mathbf{X} = \mathbf{x})$$

**Convergent message-passing algorithms.** There has been much research on finding variations to the update equations of BP that guarantee convergence. These algorithms are often similar in structure to the non-convergent algorithms, yet it can be proven that the value of the variational problem (or its dual) improves at each iteration (Hazan and Shashua 2008; Heskes 2006; Meltzer, Globerson, and Weiss 2009). Another body of recent papers have suggested to solve the convergence problems of MM-inference by linearizing the update equations. Krzakala et al. study a form of linearization for unsupervised classification called "spectral redemption" in the stochastic block model. That model is unsupervised and has no obvious way to include supervision in its setup (i.e., it is not clear how to leverage labeled nodes). Donoho, Maleki, and Montanari propose "approximate message-passing" (AMP) as an iterative thresholding algorithm for compressed sensing that is largely inspired by BP. Koutra et al. linearize BP for the case of two classes and proposed "Fast Belief Propagation" (FaBP) as a method to propagate existing knowledge of homophily or heterophily to unlabeled data. This framework allows one to specify a homophily factor $h$ ($h > 0$ for homophily or $h < 0$ for heterophily) and to then use this algorithm with exact convergence criteria for binary classification. Gatterbauer et al. derive a multivariate ("polytomous") generalization of FaBP from binary to multiple labels called "Linearized Belief Propagation" (LinBP). Both aforementioned papers show considerable speed-ups for the application of node classification and relational learning by transforming the update equations of BP into an efficient matrix formulation. However, those papers solve only special cases: FaBP is restricted to two classes per node (de facto, one single score). LinBP can handle multiple classes, but is restricted to one single node type, one single edge type, and a potential that is symmetric and doubly stochastic (see Fig. 1).[2]

---

[1]See (Murphy 2012, ch. 5.2.1) for a detailed discussion on why MAP has some undesirable properties and is not necessarily a "representative" assignment. While in theory it is arguably preferable to compute marginal probabilities, in practice researchers often use MAP inference due to the availability of efficient discrete optimization algorithms (Korč, Kolmogorov, and Lampert 2012).

**Contributions.** This paper derives a linearization of BP for *arbitrary pairwise MRFs*, which transforms the parameters of an MRF into an equation system that replaces multiplication with addition. In contrast to standard BP, the derived update equations ($i$) come with exact convergence guarantees, ($ii$) allow a closed-form solution, ($iii$) keep the derived beliefs normalized at each step, and ($iv$) can thus be put into an efficient linear algebra framework. We also show empirically that this approach – in addition to its compelling computational advantages – performs comparably to Loopy BP for a large part of the parameter space. In contrast to prior work on linearizing BP, we remove any restriction on the potentials and solve the most general case for pairwise MRFs (see Fig. 1). Since it is known that any *higher-order MRF* can be converted to a pairwise MRF (Wainwright and Jordan 2008, Appendix E.3), the approach can be also be used for higher-order potentials. Our formalism can thus model *arbitrary heterogeneous networks*; i.e., such that have directed edges or have different types of nodes.[3] This generalization is not obvious and required us to solve several new algebraic problems: ($i$) Non-symmetric potentials modulate messages differently across both directions of an edge; each direction then requires different centering points (this is particularly pronounced for non-quadratic potentials; i.e., when nodes adjacent to an edge have different numbers of classes). ($ii$) Multiplying belief vectors with non-stochastic potentials doesn't leave them stochastic; an additional normalization would then not allow a closed-form matrix formulation as before; we instead derive a "bias term" that remains constant in the update equations and thus depends only on the network structure and the potentials but not the beliefs. ($iii$) Dealing with the full heterogenous case (multiple potentials, multiple node types, and different numbers of classes among nodes) requires a considerably more general formulation. The technical report on arXiv (Gatterbauer 2015) contains the full derivation of the results presented in this paper. An efficient Python implementation is available on Github (SSLH 2015).

|  | **BP** | **FaBP** | **LinBP** | *this work* |
|---|---|---|---|---|
| # node types | arbitrary | 1 | 1 | arbitrary |
| # node classes | arbitrary | 2 | const $k$ | arbitrary |
| # edge types | arbitrary | 1 | 1 | arbitrary |
| edge symmetry | arbitrary | required | required | arbitrary |
| edge potential | arbitrary | doubly stoch. | doubly stoch. | arbitrary |
| closed form | no | yes | yes | yes |

Figure 1: The approach proposed in this paper combines the full expressiveness and generality of Loopy Belief Propagation (BP) on pairwise MRFs with the computational advantages of Fast BP (Koutra et al. 2011) and Linearized BP (Gatterbauer et al. 2015).

## 2 BP for pairwise MRFs

A MRF is a factored representation of a joint distribution over variables $\mathbf{X}$. The distribution is defined using a set of factors $\{\phi_f \mid f \in F\}$, where each $f$ is associated with the variables $\mathbf{X}_f \subset \mathbf{X}$, and $\phi_f$ is a function from the set of possible assignments of $\mathbf{X}_f$ to $\mathbb{R}^+$. The joint distribution is defined as: $\mathbb{P}(\mathbf{X} = \mathbf{x}) = \frac{1}{Z} \prod_{f \in F} \phi_f(\mathbf{x}_f)$ where $Z$ is a normalization constant known as the partition function.

An important subclass of MRFs is that of *pairwise MRFs*, representing distributions where all of the interactions between variables are limited to those between pairs. More precisely, a pairwise MRF over a graph is associated with a set of node potentials and a set of edge potentials (Koller and Friedman 2009). The overall distribution is the normalized product of all of the node and edge potentials.

We next focus on the mechanics of BP. Consider a network of $n$ nodes where each node $s$ can be any of $k_s$ possible classes (or values). A node $s$ maintains a $k_s$-dimensional belief vector where each element $j$ represents a weight proportional to the belief that this node belongs to class $j$. Let $\boldsymbol{x}_s$ be the vector of prior beliefs (also varyingly called local evidence or node potential) and $\mathbf{y}_s$ the vector of posterior (or implicit or final) beliefs at node $s$, and require that $\boldsymbol{x}_s$ and $\mathbf{y}_s$ are normalized to 1; i.e., $\sum_{j \in [k_s]} x_s(j) = \sum_{j \in [k_s]} y_s(j) = 1$. For example, a labeled node $s$ of class $i$ is represented by $x_s(j) = 1$ for $j = i$ and $x_s(j) = 0$ for $j \neq i$. Using $\mathbf{m}_{us}$ for the $k_s$-dimensional message that node $u$ sends to node $s$, we can write the BP update equations (Murphy 2012; Weiss 2000) for the belief vector of a node $s$ as:

$$y_s(j) \leftarrow \frac{1}{Z_s} x_s(j) \prod_{u \in N(s)} m_{us}(j) \qquad (1)$$

---

[2] A potential is "doubly stochastic" if all rows and columns sum up to 1. As potentials can be scaled without changing the semantics of BP, this definition also extends to any potential where the rows and columns sum to the same value.

[3] Notice that an *underlying directed network* is still modeled as an *undirected Graphical Model* (GM). For example, while the "friendship" relation on Facebook is undirected, the "follower" relation on Twitter is directed and has different implications on the two nodes adjacent to a directed "links to"-edge. Yet, the resulting GM is still undirected, but now has *asymmetric potentials*.

Here, we write $Z_s$ for a normalizer that makes the elements of $\mathbf{y}_s$ sum up to 1. Thus, the posterior belief $y_s(j)$ is computed by multiplying the prior belief $x_s(j)$ with the incoming messages $m_{us}(j)$ from all neighbors $u \in N(s)$, and then normalizing so that the beliefs in all $k_s$ classes sum to 1. In parallel, each node sends messages to each of its neighbors:

$$m_{st}(i) \leftarrow \frac{1}{Z_{st}} \sum_j \psi_{st}(j,i)\, x_s(j) \prod_{u \in N(s) \setminus t} m_{us}(j) \quad (2)$$

Here, $\psi_{st}(j,i)$ is a proportional "coupling weight" (or "compatibility," "affinity," "modulation") that indicates the relative influence of class $j$ of node $s$ on class $i$ of node $t$. Thus, the message $m_{st}(i)$ is computed by multiplying together all incoming messages at node $s$ – except the one sent by the recipient $t$ – and then passing through the $\boldsymbol{\psi}_{st}$ edge potential. Notice that we use $Z_{st}$ in Eq. (2) as a normalizer that makes the elements of $\mathbf{m}_{st}$ sum up to $k_t$ at each iteration. As pointed out by Murphy, Weiss, and Jordan; Pearl, normalizing the messages has no effect on the final beliefs; however, this intermediate normalization of messages will become crucial in our derivations. BP then repeatedly computes the above update equations for each node until the values (hopefully) converge. At iteration $r$ of the algorithm, $y_s(j)$ represents the posterior belief of $j$ conditioned on the evidence that is $r$ steps away in the network.

## 3 Linearizing BP over any pairwise MRF

This section gives a closed form description for the final beliefs after convergence of BP in *arbitrary pairwise MRFs* under a certain limit consideration of all parameters. This is a strict and non-trivial generalization of recent works (Fig. 1). The difficulty of our generalization lies in technical details: non-symmetric potentials require different centering points for messages across different directions of an edge; non-stochastic potentials require different normalizers for different iterations (and for different potentials in the networks) which does not easily lead to a simple matrix formulation; and *the full heterogenous case* (e.g., different number of classes $k$ for different nodes) requires a considerably more general derivation and final formulation.

Our approach is conceptually simple: we center all matrix entries around *well-chosen* default values and then focus only on the deviations from these defaults using Maclaurin series at several steps in our derivation. The resulting equations replace multiplication with addition and can thus be put into the framework of matrix-vector multiplication, which can leverage existing highly-optimized code. It also allows us to give exact convergence criteria for the resulting update equations and a closed form solution (that would require the inversion of a large matrix). The approach is similar in spirit to the idea of writing any MRF (with strictly positive density) as log-linear model. However, by starting from the update equations for loopy BP, we solve the intractability problem by ignoring all dependencies between messages that have traveled over a path of length 2 or more.

**Definition 1** (Centering). *We call a vector $\mathbf{x}$ or matrix $\mathbf{X}$ "centered around $c$ with standard deviation $v$" if the average entry $\mu(\mathbf{x}) = c$ and standard deviation $\sigma(\mathbf{x}) = v$.*

**Definition 2** (Residual vector/matrix). *If a vector $\mathbf{x}$ is centered around $c$, then the "residual vector" $\hat{\mathbf{x}}$ around $c$ is defined as $\hat{\mathbf{x}} = [x_1 - c, x_2 - c, \ldots]^\mathsf{T}$. Accordingly, we denote a matrix $\hat{\mathbf{X}}$ as a "residual matrix" if each entry is the residual after centering around $c$.*

For example, the vector $\mathbf{x} = [1.1, 1.2, 0.7]^\mathsf{T}$ is centered around $c = 1$, and the residuals from 1 form the residual vector $\hat{\mathbf{x}} = [0.1, 0.2, -0.3]^\mathsf{T}$; i.e., $\mathbf{x} = \mathbf{1}_3 + \hat{\mathbf{x}}$, where $\mathbf{1}_3$ is the 3-dimensional vector with all entries equal to 1. By definition of a normalized vector, beliefs for any node $s$ are centered around $\frac{1}{k_s}$, and the residuals for prior beliefs have non-zero elements (i.e., $\hat{\boldsymbol{x}}_s \neq \mathbf{0}_{k_s}$) only for nodes with local evidence (nodes "with explicit beliefs"). Further notice that the entries in a residual vector or matrix always sum up to 0 (i.e., $\sum_i \hat{x}(i) = 0$). This is done by construction and will become important in the derivations of our results.

The main idea of our derivation relies then on the following observation: if we start with messages and potentials with rows and columns centered around 1 with small enough standard deviations, then the normalizer of the update equation Eq. (2) is independent of the beliefs and *remains constant* as $Z_{st} = k_t^{-1}$. Importantly, the resulting equations *do not require further normalization*. The derivation further makes use of certain linearizing approximations that result in a well-behaved linear equation system. We show that the MM solutions implied by this equation system are *identical* to those from the original BP update equations in case of nearly uniform priors and potentials. For strong priors and potentials (e.g., $\begin{bmatrix} 1 & 100 \\ 100 & 1 \end{bmatrix}$), the resulting solutions are not identical anymore, yet serve as reasonable approximations in a wide range of problem parameters (see Section 4). WLOG, we start with potentials that are centered around 1 and then re-center the potentials before using them:[4]

**Definition 3** (Row-recentered residual matrix). *Let $\boldsymbol{\psi} \in \mathbb{R}^{\ell \times k}$ be centered around 1 and $\hat{\boldsymbol{\psi}}$ be the residual matrix around 1. Furthermore, let $\hat{r}(j) := \sum_i \hat{\psi}(j,i)$ be the sum of the residuals of row $j$. Then the "row-recentered residual matrix" $\hat{\boldsymbol{\psi}}'$ has entries $\hat{\psi}'(j,i) := \frac{1}{k}\bigl(\hat{\psi}(j,i) - \frac{\hat{r}(j)}{k}\bigr)$.*

Before we can state our main result, we need some additional notation. WLOG, let $[n]$ be the set of all nodes. For each node $s \in [n]$, let $k_s$ be the number of its possible classes. Let $\mathbf{k}_s := \frac{1}{k_s}\mathbf{1}_{k_s}$, i.e., the $k_s$-dimensional uniform stochastic column vector. Furthermore, let $k_{\text{tot}} := \sum_{s \in [n]} k_s$ be the sum of classes across nodes. To write all our resulting equations as one large equation system, we stack the individual explicit ($\hat{\boldsymbol{x}}$) and implicit ($\hat{\boldsymbol{y}}$) residual belief vectors together with the $\mathbf{k}_s$-vectors one underneath the other to form three $k_{\text{tot}}$-dimensional stacked column vectors. We also combine all row-recentered residual matrices into one large but sparse $[k_{\text{tot}} \times k_{\text{tot}}]$-square block matrix

---

[4]Without changing the joint probability distribution, every potential in a MRF can be scaled so that the average entry is 1. For example, given $\boldsymbol{\psi} = \begin{bmatrix} 4 & 6 & 5 \\ 6 & 8 & 7 \end{bmatrix}$, we scale by $\frac{1}{6}$ to get $\boldsymbol{\psi} = \begin{bmatrix} \frac{2}{3} & 1 & \frac{5}{6} \\ 1 & \frac{4}{3} & \frac{7}{6} \end{bmatrix}$, which has the identical semantics but is now centered around 1.

(notice that all entries for non-existing edges remain empty):

$$\hat{\mathbf{y}} := \begin{bmatrix} \hat{\mathbf{y}}_1 \\ \vdots \\ \hat{\mathbf{y}}_n \end{bmatrix}, \; \hat{\mathbf{x}} := \begin{bmatrix} \hat{\mathbf{x}}_1 \\ \vdots \\ \hat{\mathbf{x}}_n \end{bmatrix}, \; \mathbf{k} := \begin{bmatrix} \mathbf{k}_1 \\ \vdots \\ \mathbf{k}_n \end{bmatrix}, \; \hat{\boldsymbol{\psi}}' := \begin{bmatrix} \hat{\boldsymbol{\psi}}'_{11} & \cdots & \hat{\boldsymbol{\psi}}'_{1n} \\ \vdots & \ddots & \vdots \\ \hat{\boldsymbol{\psi}}'_{n1} & \cdots & \hat{\boldsymbol{\psi}}'_{nn} \end{bmatrix}$$

We can now state our main theorem:

**Theorem 4** (Linearizing Belief Propagation). *Let $\hat{\mathbf{y}}$, $\hat{\mathbf{x}}$, $\hat{\mathbf{k}}$, and $\hat{\boldsymbol{\psi}}'$ be the above defined residual vectors and matrix. Let $\epsilon$ be a bound on the standard deviation of all non-zero entries of $\hat{\boldsymbol{\psi}}'$ and $\hat{\mathbf{x}}$, $\sigma(\hat{\boldsymbol{\psi}}') < \epsilon$ and $\sigma(\hat{\mathbf{x}}) < \epsilon$. Let $\mathbf{y}_v^{BP}$ be the final belief assignment for any node $v$ after convergence of BP. Then, for $\lim_{\epsilon \to 0^+}$, $\arg\max_i y_v^{BP}(i) = \arg\max_i \hat{y}_v^{Lin}(i)$, where $\hat{\mathbf{y}}_v$ results from solving the following system of $k_{\text{tot}}$ linear equations in $\hat{\mathbf{y}}$:*

$$\hat{\mathbf{y}} = \underbrace{\hat{\mathbf{x}}}_{1^{st}} + \underbrace{\hat{\boldsymbol{\psi}}'^{\mathsf{T}} \mathbf{k}}_{2^{nd}} + \underbrace{\hat{\boldsymbol{\psi}}'^{\mathsf{T}} \hat{\mathbf{y}}}_{3^{rd}} - \underbrace{\hat{\boldsymbol{\psi}}'^{\mathsf{T}2} \hat{\mathbf{y}}}_{4^{th}} \quad (3)$$

In other words, the MM node labeling from BP can be approximated by solving a linear equation system if each of the potentials and each of the beliefs are reasonably tightly centered around their average values. Notice that the 2$^{nd}$ term $\hat{\boldsymbol{\psi}}'^{\mathsf{T}}\mathbf{k}$ is a "bias" vector that depends only on the structure of the network and the potentials, but *not* the beliefs. We thus sometimes prefer to write $\hat{\mathbf{c}}'_* := \hat{\boldsymbol{\psi}}'^{\mathsf{T}}\mathbf{k}$ to emphasize that it remains constant during the iterations. This term vanishes if all potentials are doubly stochastic. Also notice that the 4$^{th}$ term is what was called the "echo cancellation" in (Gatterbauer et al. 2015).[5] Simple algebraic manipulations then lead a closed-form solution by solving Eq. (3) for $\hat{\mathbf{y}}$:

$$\hat{\mathbf{y}} = \left(\mathbf{I}_{k_{\text{tot}}} - \hat{\boldsymbol{\psi}}'^{\mathsf{T}} + \hat{\boldsymbol{\psi}}'^{\mathsf{T}2}\right)^{-1}\left(\hat{\mathbf{x}} + \hat{\mathbf{c}}'_*\right) \quad (4)$$

### Iterative updates and convergence

The complexity of inverting a matrix is cubic in the number of variables, which makes direct application of Eq. (4) difficult. Instead, we use Eq. (3), which gives an implicit definition of the final beliefs, iteratively. Starting with an arbitrary initialization of $\hat{\mathbf{y}}$ (e.g., all values zero), we repeatedly compute the right hand side of the equations and update the values of $\hat{\mathbf{y}}$ until the process converges:[6]

---

[5]Notice that the BP update equations send a message across an edge that excludes information received across the same edge from the other direction: "$u \in N(s) \setminus t$" in Eq. (2). In a probabilistic scenario on tree-based graphs, this echo cancellation is required for correctness. In loopy graphs (without well-justified semantics), this term still compensates for the message a node $t$ would otherwise send to itself via a neighbor $s$, i.e., via the path $t \to s \to t$.

[6]Interestingly, our linearized update equations, Eq. (5), are reminiscent of the update equations for the mean beliefs in Gaussian MRFs (Malioutov, Johnson, and Willsky 2006; Su and Wu 2015; Weiss and Freeman 2001). Notice however, that whereas the update equations are exact in the case of continuous Gaussian MRFs, our equations are approximations for the general discrete case.

**Proposition 5** (Update equations). *The positive fix points for Eq. (3) can be calculated iteratively with the following update equations starting from $\hat{\mathbf{y}}^{(0)} = \mathbf{0}$:*

$$\hat{\mathbf{y}}^{(r+1)} \leftarrow (\hat{\mathbf{x}} + \hat{\mathbf{c}}'_*) + (\hat{\boldsymbol{\psi}}'^{\mathsf{T}} - \hat{\boldsymbol{\psi}}'^{\mathsf{T}2})\hat{\mathbf{y}}^{(r)} \quad (5)$$

These particular update equations allow us to give a sufficient and necessary criterium for convergence via the spectral radius $\rho$ of a matrix.[7]

**Corollary 6** (Convergence). *The update Eq. (5) converges if and only if $\rho(\hat{\boldsymbol{\psi}}' - \hat{\boldsymbol{\psi}}'^2) < 1$.*

Thus, the updates converge towards the closed-form solution, and the final beliefs of each node can be computed via efficient matrix operations with optimized packages, while the implicit form gives us guarantees for the convergence of this process.[8] In order to apply our approach to problem settings with spectral radius bigger than one (and thus direct application of Eq. (5) would not work), we propose to modify the model by *weakening the potentials*. In other words, we multiply $\hat{\boldsymbol{\psi}}'$ with a factor that guarantees convergence. We call the multiplicative factor which exactly separates convergence from divergence, the "*convergence boundary*" $\epsilon_*$. Choosing any $\epsilon$ with $s := \frac{\epsilon}{\epsilon_*}$ and $s < 1$ guarantees convergence. We call any choice of $s$ the "*convergence parameter.*"

**Definition 7** (Convergence boundary $\epsilon_*$). *For any $\hat{\boldsymbol{\psi}}'$, the convergence boundary $\epsilon_* > 0$ is defined implicitly by $\rho(\epsilon_* \hat{\boldsymbol{\psi}}' - \epsilon_*^2 \hat{\boldsymbol{\psi}}'^2) = 1$.*

### Computational complexity

Naively materializing $\hat{\boldsymbol{\psi}}'$ would lead to a space requirement of $\mathcal{O}(n^2 k_{\max}^2)$ where $n$ is the number of nodes and $k_{\max}$ the max number of classes per node. However, by using a sparse matrix implementation, both the space requirement and the computational complexity of each iteration are only proportional to the number of edges: $\mathcal{O}(m k_{\max}^2)$. The time complexity is identical to the one of message-passing *with division*, which avoids redundant calculations and is faster than standard BP on graphs with high node degrees (Koller and Friedman 2009). However, the ability to use existing highly-optimized packages for efficient matrix-vector multiplication will considerably speed-up the actual calculations.

## 4 Experiments

**Questions.** Our experiments will answer the following 3 questions: (1) What is the effect of the convergence parameter $s$ on accuracy and number of required iterations until convergence? (2) How accurate is our approximation under varying conditions: ($i$) the density of the network, ($ii$) the strength on the interaction, and ($iii$) the fraction of labeled nodes? (3) How fast is the linearized approximation as compared to standard Loopy BP?

---

[7]The "spectral radius" $\rho(\cdot)$ of a matrix is the supremum among the absolute values of its eigenvalues.

[8]The intuition behind these equivalences can be illustrated by comparing to the geometric series $S = 1 + x + x^2 + \ldots$ and its closed form $S = (1-x)^{-1}$. Whereas for $|x| < 1$, the series converges to its closed-form, for $|x| > 1$, it diverges, and the closed-form is meaningless.

**Experimental protocol.** We define "accuracy" as the fraction of unlabeled nodes that receive correct labels. In order to evaluate the accuracy of a method, we need to use graphs with known label ground truth (GT). As we are interested in the accuracy as a function of various parameters, we need graphs with *controlled GT*. We thus decided to compare BP against its linearization on synthetic graphs with known GT, which allows us to measure the accuracy as result of *systematic parameter changes*. The well-studied stochastic block-model (Airoldi et al. 2008) leads to networks with degree distributions that are not similar to those found in most empirical network data. Our synthetic graph generator is thus a variant thereof with two important differences: (1) we actively control the degree distributions in the resulting graph; and (2) we "plant" exact graph properties (instead of fixing a property only in expectation). In other words, our generator preserves desired degree distribution and compatibilities between classes. The online appendix (Gatterbauer 2015) contains all details. We focus on the scenario of a network with one non-symmetric potential along each edge. The generator creates a graph using a tuple of parameters $(n, m, \boldsymbol{\alpha}, \boldsymbol{\psi}, \mathsf{dist})$, where $n$ is the number of nodes, $m$ is the number of edges, $\boldsymbol{\alpha}$ is the node label distribution with $\alpha(i)$ being the fraction of nodes of class $i$, $\boldsymbol{\psi}$ is the edge potential, and $\mathsf{dist}$ is a chosen degree distribution (e.g., uniform or power law with chosen coefficient).

**Parameter choices.** Throughout our experiments, we use $k = 3$ classes and the potential $\boldsymbol{\psi} = \begin{bmatrix} 1 & h & 1 \\ 1 & 1 & h \\ h & 1 & 1 \end{bmatrix}$, parameterized by a value $h$ representing the ratio between min and max entries. Dividing by $(2 + h)$ centers it around 1. Thus parameter $h$ models the *strength of the potential*, and we expect higher values of $h$ to make our approximation less suitable. Notice that this matrix is not symmetric and shows very different modulation behavior across both directions of an edge. We create graphs with $n$ nodes and assign the same fraction of nodes to one of the 3 classes: $\boldsymbol{\alpha} = [\frac{1}{3}, \frac{1}{3}, \frac{1}{3}]$. We also vary the parameters $m$ and $d = \frac{m}{n}$ as the average in- and outdegree in the graph, and we assume a power law distribution with coefficient $0.5$. We then keep a fraction $f$ of node labels and measure accuracy on the remainder.

**Computational setup.** All methods are implemented in Python and use the optimized SciPy library (Jones et al. 2001) to handle sparse matrix operations. The experiments are run on a 2.5 Ghz Intel Core i5 with 16G of main memory and a 1TB SSD hard drive. To allow comparability across implementations, we limit evaluation to one processor. For timing BP, we use message-passing with division which is faster than standard BP on graphs with high node degree (Koller and Friedman 2009). To calculate the approximate spectral radius of a matrix, we use a method from the PyAMG library (Bell, Olson, and Schroder 2011) that implements a technique described in (Bai et al. 2000). Our code, including the data generator, is inspired by Scikit-learn (Pedregosa et al. 2011) and is available on Github to encourage reproducible research (SSLH 2015).

> **Question 1.** *What is the effect of scaling parameter $s$ on accuracy and number of iterations for convergence?*
>
> *Result 1.* Our scaling parameter $s$ gives an exact criterion for our approach to converge. In contrast, BP often does not converge and requires a lot of fine-tuning; e.g., damping or even scaling of the potential. The accuracy of the linearization is highest for $s$ close or slightly above 1 and by not iterating until convergence.

Figure 2a shows the number of required iterations to reach convergence and confirms our theoretical results from Corollary 6. In case the convergence condition does not hold, we scale the centered potential by a value $\epsilon$, resulting from $\epsilon = s \cdot \epsilon_*$ with $s < 1$. This action weakens the potentials, but preserves the relative affinities (we also use the same approach to help BP find a fixed point if it does not converge within 200 iterations). Figure 2b shows what happens to accuracy if we run the iterative updates a fixed number of times as a function of $s$. Notice that even *considerably* scaling a potential does not entirely change the model and still gives reasonable approximations. The figure fixes a number of iterations, but then varies again $\epsilon$ via $s$. Also interestingly, almost all of the performance gains from the linearized update equations come from running just a few iterations, and convergence for optimal labeling is not necessary; instead, by choosing $s \approx 1$ (at the *exact boundary of convergence*) or even $s > 1$ and iterating only a few times, we can maximize the expected accuracy. For the remaining accuracy experiments, we use $s = 0.5$ and run our algorithm to convergence.

> **Question 2.** *How accurate is our approximation, and under which conditions is it reasonable?*
>
> *Result 2.* The linearization gives comparable labeling accuracy as LBP for graphs with weak potentials. The performance deteriorates the most in *dense* networks with *strong potentials*.

We found that $h$, $d$ and $f$ have important influence on the labeling accuracy of BP and its linearization (whereas $n$, $\mathsf{dist}$ and $\boldsymbol{\alpha}$ influence only to a lesser extent). Figures 2c and 2d show accuracy as a function of the fraction $f$ of labeled nodes. Notice that we chose the best BP was able to perform (over several choices of $\epsilon$ and damping factors to make it converge) whereas for LinBP we consistently chose $s = 0.5$ as proposed in (Gatterbauer et al. 2015). Figures 2e to 2g show labeling quality as a function the strength $h$ of the potential. For strong potentials ($h > 3$), BP gives better accuracy *if* it converges. In practice, BP often did not converge within 200 iterations even for weak potentials (bordered data points required dampening; red crosses required additional entry-wise scaling of the potential with our convergence boundary $\epsilon_*$). In our experiments, BP often did not converge despite using damping, surprisingly often when $h$ is not big. It is known that if the potentials are close to indifference then loopy BP usually converges. In this case, our formalism is equivalent to loopy BP (this follows from our linearization). Thus, whenever loopy BP did not converge, we simply exponentiated the entries of the potential with a varying factor $\epsilon$ until BP converged. Thus for high $h$, BP *can* perform better than the linearization, but only after a lot of fine-tuning of parameters. In contrast, for our formulation we know exactly the boundary of convergence.

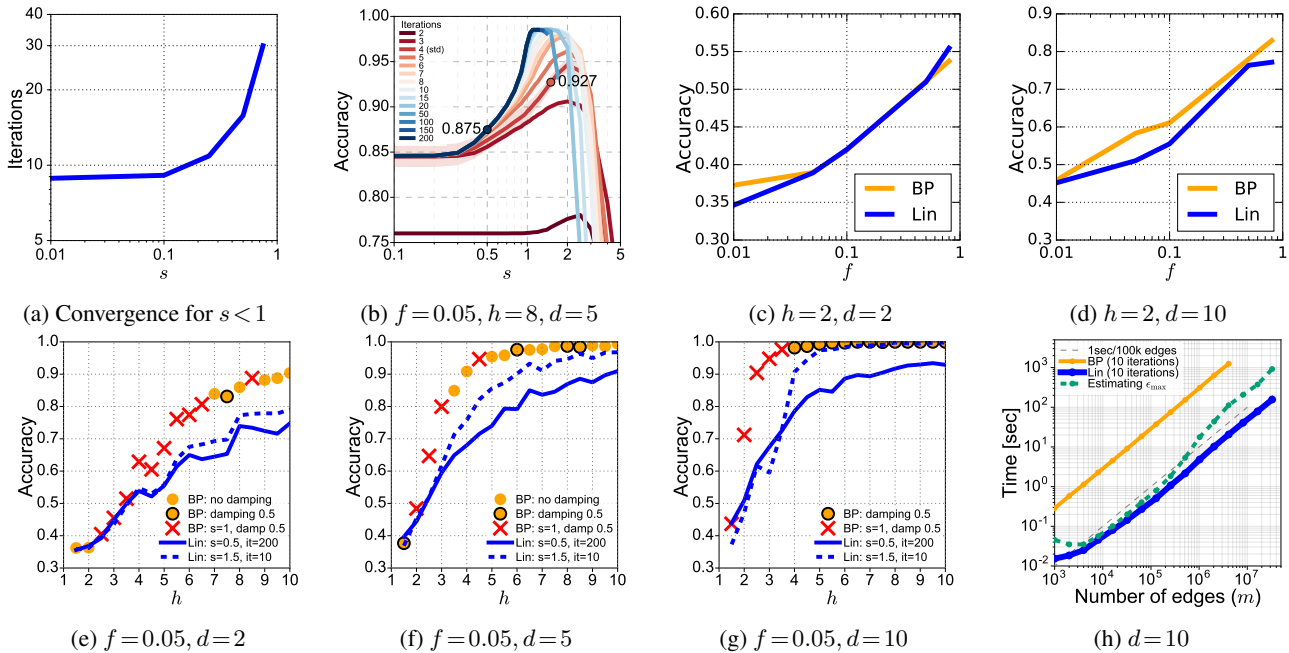

Figure 2: Experimental results for BP and its linearization (abbreviated here as "Lin"): $f$ represents the fraction of labeled nodes, $h$ the strength of potentials, and $d$ the average node in- and outdegree. All graphs except for (h) have $n=1000$ nodes. (a): The *convergence parameter $s$* exactly determines convergence for the linearization. (b): Accuracy increases for $s$ close to 1 and *few iterations*. (c, d): $f$ only marginally affects the relative accuracy between BP and its linearization. (e, f, g): For strong potentials, BP gives better accuracy *if it converges*. In practice, BP often did not converge within 200 iterations even for weak potentials and required a lot of fine-tuning (damping and/or entry-wise scaling of the potential with our *convergence boundary* $\epsilon_*$ to $s=1$). (h): Each iteration of our approach is 50 times faster than an implementation of BP with division. In addition, deploying a proper damping strategy often requires 100s of iterations, which can bring up the total speed-up to a factor 1000 for some of the above data points. (Each data point results from at least 10 samples).

Overall, the linearization gives comparable results to the original BP for small potentials, and BP performance is better than the linearization only either for strong potentials with $h \geq 3$ and dampening (see a few yellow dots without borders as exceptions) or after fine-tuning BP after using our own convergence boundary and scaling the potentials, or after a lot of manual fine-tuning.

> **Question 3.** *How fast is the linearized approximation as compared to BP?*
>
> *Result 3.* The linearization is around 100 times faster than BP per iteration and often needs 10 times fewer iterations until convergence. In practice, this can lead to a speed-up of 1000 times.

A key advantage of the linearization is that it has *predictable* convergence and comes with considerable speed-ups. Figure 2h shows that our approach scales linearly in the number of edges and is 50 times faster than regular loopy BP *per iteration*; an iteration on a graph with 3 million nodes and 30 million edges takes less than 2 sec. Calculating the exact convergence boundary via a spectral radius calcula-tion can take more time (approx. 1000 sec for the same graph). Notice that any dampening strategy for BP results in increased number of iterations and needs to overcome the additional slow-down of further iterations. Also recall that on each circled point in Figs. 2e to 2g, BP did *not converge* within 200 iterations and required dampening; each red cross required *additional scaling* of the potentials with our calculated $\epsilon_*$ in order to make BP converge.

## 5 Conclusions

We have derived a linearization of BP for arbitrary pairwise MRFs for the purpose of node labeling with MM-inference. The approach transforms the parameters of an MRF into a linear equation system that can be solved with simple iterative updates. These updates come with exact convergence guarantees, allow a closed-form solution, keep the derived beliefs normalized at each step, and can thus be put into an efficient linear algebra framework that does not require normalization at each step. Experiments on carefully controlled synthetic data with known ground truth show that our approach performs comparably with Loopy BP for weak potentials and comes with a predictable behavior, compelling computational advantages, and an easy implementation with only few lines of code. An unexplored application of the linearization may be speeding-up convergence of regular BP by starting from good approximations of its fixed points.

**Acknowledgements.** This work was supported in part by NSF grant IIS-1553547. I would like to thank Christos Faloutsos for very convincingly persuading me of the power of linear algebra and continued support. I am also grateful to Stephan Günneman, Vladimir Kolmogorov, and Christoph Lampert for a number of insightful comments.

## A  Nomenclature

| | |
|---|---|
| $n$ | number of nodes |
| $m$ | number of edges |
| $s, t, u$ | indices used for nodes |
| $N(s)$ | set of neighbors of node $s$ |
| $k, \ell$ | number of classes, $k_s$ is number of classes for node $s$, $k(s)$ is the class of node $s$ |
| $i, j, g$ | indices used for classes |
| $(r)$ | index used for iteration |
| $\boldsymbol{x}_s$ | $k_s$-dimensional prior (or explicit) belief vector of node $s$ |
| $\mathbf{y}_s$ | $k_s$-dimensional posterior (or implicit or final) belief vector of node $s$ |
| $\mathbf{m}_{st}$ | $k_t$-dimensional message vector from node $s$ to node $t$ ($s \to t$) |
| $\boldsymbol{\psi}$ | $\ell \times k$ potential (or coupling matrix): $\psi(j, i)$ indicates the influence of class $j$ of the sending node on class $i$ of the receiving node. WLOG potentials are scaled to be centered around 1. |
| $\hat{\mathbf{y}}, \hat{y}(j)$ | The hat notation "ˆ" indicates residuals after centering |
| $\hat{\boldsymbol{\psi}}', \hat{\boldsymbol{\psi}}''$ | row-recentered " ′ " or column-recentered " ″ " residual potential |
| $\hat{\mathcal{P}}'$ | set of all row-recentered residual potentials |
| $Q$ | set of node types |
| $q$ | index used for node types |
| $k(q)$ | number of classes for node type $q$ ($q \in Q$) |
| $K$ | set of numbers of classes across all nodes: $K = \{k(1), k(2), \ldots k(|Q|)\}$ |
| $N_k$ | set of nodes with $k$ classes |
| $n_k$ | number of nodes with $k$ classes |
| $o(s)$ | order (sequence) of node $s$ within $N_{k_s}$ |
| $\mathbf{X}_k, \mathbf{Y}_k$ | $n \times k$ prior or posterior belief matrix: $X(o(s), j)$ indicates the belief in class $j$ by node $s$ |
| $\mathbf{W}_{\boldsymbol{\psi}}$ | $n_\ell \times n_k$ adjacency matrix for edges with potential $\boldsymbol{\psi} \in \mathbb{R}^{\ell \times k}$: $W(o(s), o(t)) \neq 0$ indicates an edge $s \to t$ that carries the potential $\boldsymbol{\psi}$ |
| $\mathbf{I}_k$ | $k \times k$ identity matrix |
| $\mathbf{X}^\intercal$ | transpose of matrix $\mathbf{X}$ |
| $\text{vec}(\mathbf{X})$ | vectorization of matrix $\mathbf{X}$ |
| $\mathbf{X} \otimes \mathbf{Y}$ | Kronecker product between matrices $\mathbf{X}$ and $\mathbf{Y}$ |
| $\mathbf{X} \odot \mathbf{Y}$ | Hadamard product or component-wise multiplication: $\mathbf{Z} = \mathbf{X} \odot \mathbf{Y} \Leftrightarrow Z(i,j) = X(i,j) \cdot Y(i,j)$ |
| $\mathbf{X} \oslash \mathbf{Y}$ | Component-wise division: $\mathbf{Z} = \mathbf{X} \oslash \mathbf{Y} \Leftrightarrow Z(i,j) = X(i,j)/Y(i,j)$ with $0/0 = 0$ |
| $\frac{1}{Z}$ | normalizer |
| $\mathbf{1}_k$ | $k$-dimensional column vector with all entries equal to 1 |
| $[x]_{\ell \times k}$ | $\ell \times k$ matrix with all entries equal to $x$ |
| $\boldsymbol{\alpha}$ | $k$-dimensional node label distribution |
| $[k]$ | $[k] := \{1, 2, \ldots, k\}$ |

Figure 3: Nomenclature

Given a matrix $\mathbf{X}$, we write $X(i, j)$ for one scalar entry, $\mathbf{X}(i, :)$ for the $i$-th row vector, and $\mathbf{X}(:, j)$ for the $j$-th column vector. We also write $\sum_j$ as short form for $\sum_{j \in [k]}$ whenever $k$ is clear from the context.

## B  Derivation of the linearization of BP over any pairwise MRFs

This section contains the derivation of Theorem 4. We will center the elements of all message and belief vectors around their "natural default values," i.e., the elements of $\mathbf{m}_{st}$ around 1, and the elements of $\boldsymbol{x}_s$ and $\mathbf{y}_s$ around $\frac{1}{k_s}$ (Lemma 10 will provide some intuition why our chosen center points are the natural choice to simplify all later derivations). We are interested in the residual values defined by $\hat{m}(i) := m(i) - 1$, $\hat{x}_s(j) := x_s(j) - \frac{1}{k_s}$, and $\hat{y}_s(j) := y_s(j) - \frac{1}{k_s}$.

WLOG, we start from a potential $\boldsymbol{\psi} \in \mathbb{R}^{\ell \times k}$ that is centered around 1 (Recall that we can scale any potential with a positive real number without changing the semantics of the MRF). We then appropriately *recenter* a potential differently across both directions of an edge as to make it singly stochastic for either direction and most of the residual terms for the belief update equations cancel each other out, leading to simplified equations. Definition 3 provided the definition for the residual matrix in one direction, *row-recentering*. Adding to that definition, the *row-recentered stochastic matrix* $\boldsymbol{\psi}'$ is centered around $\frac{1}{k}$ and has entries $\psi'(j, i) := \hat{\psi}'(j, i) + \frac{1}{k}$. Both matrices are indicated with a single apostrophe ′.

Analogously, let $\hat{c}(i) = \sum_i \hat{\psi}(j, i)$ be the residual sum of column $i$. Then, a *column-recentered residual matrix* $\hat{\boldsymbol{\psi}}''$ has entries $\hat{\psi}''(j, i) := \frac{1}{\ell}(\hat{\psi}(j, i) - \frac{\hat{c}(i)}{\ell})$ and the *column-recentered stochastic matrix* $\boldsymbol{\psi}''$ has entries $\psi''(j, i) := \frac{1}{\ell} + \hat{\psi}''(j, i)$. Notice that both matrices are indicated with a double apostrophe ″. The resulting recentered residual potentials are coupling matrices that make explicit the relative attraction and repulsion of neighboring nodes. For example, the sign of $\hat{\psi}'(j, i)$ tells us if the class $j$ attracts or repels class $i$ in a neighbor, and the magnitude of $\hat{\psi}'(j,i)$ indicates the extent. Subsequently, this centering allows us to rewrite belief propagation in terms of the residuals.

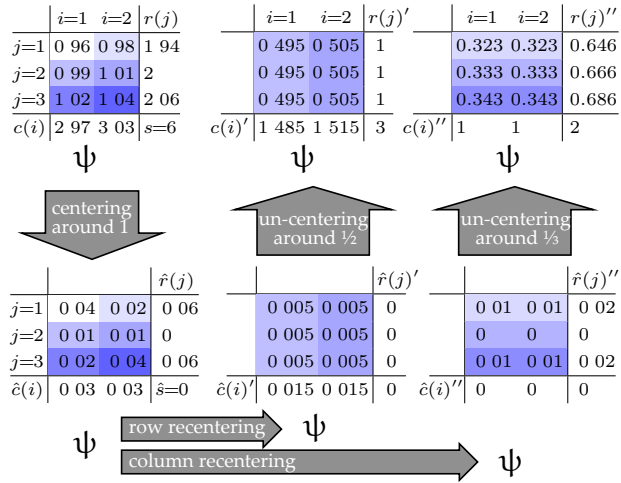

Figure 4: Example 9: Matrix $\boldsymbol{\psi} \in \mathbb{R}^{\ell \times k}$ centered around 1, residual matrix $\hat{\boldsymbol{\psi}}$, and row-recentered and column-recentered residual matrices $\hat{\boldsymbol{\psi}}'$ and $\hat{\boldsymbol{\psi}}''$ (and stochastic matrices $\boldsymbol{\psi}'$ and $\boldsymbol{\psi}''$).

|  | Expression/Maclaurin series/Approximation |
|---|---|
| Logarithm | $\ln(1+\epsilon) = \epsilon - \frac{\epsilon^2}{2} + \frac{\epsilon^3}{3} - \ldots \approx \epsilon$ |
| Product | $(1+\epsilon_1)(1+\epsilon_2) = 1+\epsilon_1+\epsilon_2+\epsilon_1\epsilon_2 \approx 1+\epsilon_1+\epsilon_2$ |
| Division | $\frac{1/k+\epsilon_1}{1+\epsilon_2} = (\frac{1}{k}+\epsilon_1)(1-\epsilon_2+\epsilon_2^2-\ldots) \approx \frac{1}{k}+\epsilon_1-\frac{\epsilon_2}{k}$ |
| Scaling | $(1-\epsilon^2)^{-1} \approx 1$ |

Figure 5: Table of our linearizing approximations.

Notice that column-recentering and row-recentering are connected via the transpose. However, message modulation across one direction of an edge is is *not simply the transpose* of the modulation across the other direction:

**Corollary 8** (Row-recentering vs. column-recentering). $(\hat{\boldsymbol{\psi}}'')^\mathsf{T} = (\hat{\boldsymbol{\psi}}^\mathsf{T})'$. *In particular,*
$$\hat{\boldsymbol{\psi}}''_{st} = (\hat{\boldsymbol{\psi}}'_{ts})^\mathsf{T}$$

We also write the $\ell$-dimensional vector $\mathbf{r} := \boldsymbol{\psi} \mathbf{1}_k$ for the row sums, the $k$-dimensional vector $\mathbf{c} := \boldsymbol{\psi}^\mathsf{T} \mathbf{1}_\ell$ for the column sums, and $s := \sum_j r(j) = \mathbf{1}_\ell^\mathsf{T} \mathbf{r} = \mathbf{1}_\ell^\mathsf{T} \boldsymbol{\psi} \mathbf{1}_k$ for the sum of all entries in a matrix. We illustrate recentering next with a detailed example.

**Example 9** (Recentering). *Figure 4 shows the $3 \times 2$ matrix $\boldsymbol{\psi}$ that is centered around 1 (i.e., each entry is close to 1 and the average is exactly 1) together with the row sums $r(j)$ and the column sums $c(i)$. $\hat{\boldsymbol{\psi}}$ is then the residual matrix. Notice that the recentered residual matrices $\hat{\boldsymbol{\psi}}'$ and $\hat{\boldsymbol{\psi}}''$ have zero row sums $\hat{r}(j)'$ or column sums $\hat{c}(i)''$, respectively. As consequence, the row-recentered matrix $\boldsymbol{\psi}'$ and column-recentered matrix $\boldsymbol{\psi}''$ are* row-stochastic *or* column-stochastic, *respectively.*

We will further make use of the linearizing Maclaurin series approximations shown in Fig. 5 to derive a well-behaved linear equation system.

**Recentering**

The following lemma provides the mathematical justification for our particular choice of recentering:

**Lemma 10** (Recentering). *Consider the update equation*

$$\mathbf{y} \leftarrow \frac{1}{Z} \boldsymbol{\psi}^\mathsf{T} \mathbf{x} \tag{6}$$

*with $\mathbf{x}$ being a $\ell$-dimensional stochastic vector, $\boldsymbol{\psi} \in \mathbb{R}^{\ell \times k}$ being centered around 1, and $Z$ a normalizer that makes the elements of the resulting $k$-dimensional vector $\mathbf{y}$ sum up to $k$. Then, the update equation can be approximated with the row-recentered stochastic matrix $\boldsymbol{\psi}'$ by*

$$\mathbf{y} \leftarrow k\,\boldsymbol{\psi}'^\mathsf{T} \mathbf{x} \tag{7}$$

*Proof Lemma 10.* Our proof will express both equations (Eq. (6) with $\boldsymbol{\psi}$ and Eq. (7) with $\boldsymbol{\psi}'$) in terms of the residual matrix $\hat{\boldsymbol{\psi}}'$, and show that they lead to the same equation. From Definition 3 and the definitions at the beginning of Appendix B, we know that $\psi(j,i) = 1 + \hat{\psi}(j,i)$ and $\hat{\psi}(j,i) = k\,\hat{\psi}'(j,i) + \frac{\hat{r}(j)}{k}$. Therefore, $\psi(j,i) = 1 + k\,\hat{\psi}'(j,i) + \frac{\hat{r}(j)}{k}$. Similarly, $\psi'(j,i) = \frac{1}{k} + \hat{\psi}'(j,i)$.

In the following, we are going to use matrix notation that allows us to express the above identities very compactly as: $\boldsymbol{\psi}^\mathsf{T} = \mathbf{1}_k \mathbf{1}_\ell^\mathsf{T} + \frac{1}{k}\mathbf{1}_k \hat{\mathbf{r}}^\mathsf{T} + k\,\hat{\boldsymbol{\psi}}'^\mathsf{T}$, $\boldsymbol{\psi}'^\mathsf{T} = \frac{1}{k}\mathbf{1}_k \mathbf{1}_\ell^\mathsf{T} + \hat{\boldsymbol{\psi}}'^\mathsf{T}$, and $\mathbf{x} = \frac{1}{\ell}\mathbf{1}_\ell + \hat{\mathbf{x}}$.[9]

($i$) Equation (6): We calculate $\mathbf{y}$ in two steps that treat the normalization separately: first $\mathbf{z} = \boldsymbol{\psi}^\mathsf{T}\mathbf{x}$, and then $\mathbf{y} = \frac{1}{Z}\mathbf{z}$.

$$\mathbf{z} = \boldsymbol{\psi}^\mathsf{T}\mathbf{x}$$
$$= \left(\mathbf{1}_k \mathbf{1}_\ell^\mathsf{T} + \frac{1}{k}\mathbf{1}_k \hat{\mathbf{r}}^\mathsf{T} + k\hat{\boldsymbol{\psi}}'^\mathsf{T}\right) \cdot \left(\frac{1}{\ell}\mathbf{1}_\ell + \hat{\mathbf{x}}\right)$$
$$= \mathbf{1}_k + \frac{1}{k\ell}\mathbf{1}_k \underbrace{\hat{s}}_{=0} + \frac{k}{\ell}\underbrace{\hat{\boldsymbol{\psi}}'^\mathsf{T}\mathbf{1}_\ell}_{=\hat{\mathbf{c}}'} + \mathbf{1}_k \underbrace{\mathbf{1}_\ell^\mathsf{T}\hat{\mathbf{x}}}_{=0} + \frac{1}{k}\mathbf{1}_k \hat{\mathbf{r}}^\mathsf{T}\hat{\mathbf{x}} + k\hat{\boldsymbol{\psi}}'^\mathsf{T}\hat{\mathbf{x}}$$
$$= \mathbf{1}_k + \frac{1}{k}\mathbf{1}_k \hat{\mathbf{r}}^\mathsf{T}\hat{\mathbf{x}} + \frac{k}{\ell}\hat{\mathbf{c}}' + k\hat{\boldsymbol{\psi}}'^\mathsf{T}\hat{\mathbf{x}}$$

We next calculate the value of the normalizer. Recall that the normalizer makes the entries of the vector $\mathbf{z}$ sum up to $k$.

$$Z = \frac{1}{k}\mathbf{1}_k^\mathsf{T}\mathbf{z} = \frac{1}{k}\left(k + \hat{\mathbf{r}}^\mathsf{T}\hat{\mathbf{x}} + \frac{k}{\ell}\underbrace{\mathbf{1}_k^\mathsf{T}\hat{\mathbf{c}}'}_{=0} + k\underbrace{\mathbf{1}_k^\mathsf{T}\hat{\boldsymbol{\psi}}'^\mathsf{T}\hat{\mathbf{x}}}_{=0}\right)$$
$$= 1 + \frac{\hat{\mathbf{r}}^\mathsf{T}\hat{\mathbf{x}}}{k}$$

We see that the normalizer is not a constant but also depends on $\boldsymbol{\psi}$ and $\mathbf{x}$. However, notice that if each row of $\boldsymbol{\psi}$ is centered around 1 (not just the matrix as a whole), then $\hat{r}(j) = 0$ for all rows, and thus $Z = 1$. In the following, we approximate $1/(1+\epsilon) \approx (1-\epsilon)$ and $(1+\epsilon_1)(1+\epsilon_2) \approx 1 + \epsilon_1 - \epsilon_2$.

$$\mathbf{y} = \left(\mathbf{1}_k + \frac{1}{k}\mathbf{1}_k \hat{\mathbf{r}}^\mathsf{T}\hat{\mathbf{x}} + \frac{k}{\ell}\hat{\mathbf{c}}' + k\hat{\boldsymbol{\psi}}'^\mathsf{T}\hat{\mathbf{x}}\right)\left(1 + \frac{\hat{\mathbf{r}}^\mathsf{T}\hat{\mathbf{x}}}{k}\right)^{-1}$$
$$\approx \left(\mathbf{1}_k + \frac{1}{k}\mathbf{1}_k \hat{\mathbf{r}}^\mathsf{T}\hat{\mathbf{x}} + \frac{k}{\ell}\hat{\mathbf{c}}' + k\hat{\boldsymbol{\psi}}'^\mathsf{T}\hat{\mathbf{x}}\right)\left(1 - \frac{\hat{\mathbf{r}}^\mathsf{T}\hat{\mathbf{x}}}{k}\right)$$
$$\approx \mathbf{1}_k + \cancel{\frac{1}{k}\mathbf{1}_k\hat{\mathbf{r}}^\mathsf{T}\hat{\mathbf{x}}} + \frac{k}{\ell}\hat{\mathbf{c}}' + k\hat{\boldsymbol{\psi}}'^\mathsf{T}\hat{\mathbf{x}} - \cancel{\frac{1}{k}\mathbf{1}_k\hat{\mathbf{r}}^\mathsf{T}\hat{\mathbf{x}}}$$
$$= \mathbf{1}_k + \frac{k}{\ell}\hat{\mathbf{c}}' + k\hat{\boldsymbol{\psi}}'^\mathsf{T}\hat{\mathbf{x}}$$

Notice that the above equation is exact if $\hat{r}(j) = 0$ for all rows.

($ii$) Equation (7): Here we get the same result much faster:

$$\mathbf{y} = k\boldsymbol{\psi}'^\mathsf{T}\mathbf{x}$$
$$= \left(\mathbf{1}_k\mathbf{1}_\ell^\mathsf{T} + k\hat{\boldsymbol{\psi}}'^\mathsf{T}\right)\cdot\left(\frac{1}{\ell}\mathbf{1}_\ell + \hat{\mathbf{x}}\right)$$
$$= \mathbf{1}_k + k\underbrace{\hat{\boldsymbol{\psi}}'^\mathsf{T}\mathbf{1}_\ell}_{\hat{\mathbf{c}}'}\frac{1}{\ell} + \mathbf{1}_k\underbrace{\mathbf{1}_\ell^\mathsf{T}\hat{\mathbf{x}}}_{0} + k\hat{\boldsymbol{\psi}}'^\mathsf{T}\hat{\mathbf{x}}$$
$$= \mathbf{1}_k + \frac{k}{\ell}\hat{\mathbf{c}}' + k\hat{\boldsymbol{\psi}}'^\mathsf{T}\hat{\mathbf{x}}$$

---
[9] A quick illustration: $\frac{1}{k}\mathbf{1}_k\hat{\mathbf{r}}^\mathsf{T} = \frac{1}{1}[1,1]^\mathsf{T}[-0.06, 0, 0.06] = \frac{1}{2}\begin{bmatrix}-0.06 & 0 & 0.06 \\ -0.06 & 0 & 0.06\end{bmatrix}$ for $\boldsymbol{\psi}$ in Fig. 4.

It follows that Eq. (7) is an approximation of Eq. (6), in general, and both equations are equivalent if each row in $\boldsymbol{\psi}$ is centered around 1. □

Notice that, since $y(j) = 1 + \hat{y}(j)$, we can express the update equation in terms of residuals as

$$\hat{\mathbf{y}} = \frac{k}{\ell}\hat{\mathbf{c}}' + k\hat{\boldsymbol{\psi}}'^{\mathsf{T}}\hat{\mathbf{x}}$$

Further notice that if each column in the original potential is centered around 1, then the term $\hat{\mathbf{c}}'$ disappears.

Overall, Lemma 10 implies that by recentering the coupling matrix, we can replace the normalizer with a constant, which considerably simplifies our later derivations. The proof also showed that the approximation becomes exact *if each row* in $\boldsymbol{\psi}$ is centered around 1.

> **Example 11** (Recentering (Example 9 continued)). *Consider again matrix* $\boldsymbol{\psi} \in \mathbb{R}^{3\times 2}$ *in Fig. 4: The matrix is centered around 1 as the sum of its entries is $s = 6$. However, row 1 is not centered around 1 as its row sum $r(1) = 1.94$ instead of 2. Next assume $\mathbf{x} = [0.1, 0.1, 0.8]^{\mathsf{T}}$. Then $\mathbf{y} = [0.99021, 1.00979]^{\mathsf{T}}$ for Eq. (6), but $\mathbf{y} = [0.99, 1.01]^{\mathsf{T}}$ with Eq. (7). Thus, the residuals are $\pm 0.00979$ and $\pm 0.01$, respectively, and the relative difference $\approx 2\%$.*

### Centered BP

By using the previous lemma and focusing on the residuals only, we can next transform the belief update equations from multiplication into addition:

> **Lemma 12** (Centered BP). *By appropriately centering the coupling matrix, beliefs and messages, the equations for belief propagation can be approximated by:*
>
> $$\hat{y}_s(j) \leftarrow \hat{x}_s(j) + \frac{1}{k_s}\sum_{u\in N(s)} \hat{m}_{us}(j) \tag{8}$$
>
> $$\hat{m}_{st}(i) \leftarrow \frac{k_t}{k_s}\hat{c}_{st}(i)' + k_t\sum_j \hat{\psi}'_{st}(j,i)\Big(\hat{y}_s(j) - \frac{1}{k_s}\hat{m}_{ts}(j)\Big) \tag{9}$$

*Proof Lemma 12.* ($i$) Equation (8): Substituting the expansions into the belief updates Eq. (1) leads to

$$\frac{1}{k_s} + \hat{y}_s(j) \leftarrow \frac{1}{Z_s} \cdot \Big(\frac{1}{k_s} + \hat{x}_s(j)\Big) \cdot \prod_{u\in N(s)}\big(1 + \hat{m}_{us}(j)\big)$$

$$\ln\big(1 + k_s\hat{y}_s(i)\big) \leftarrow -\ln Z_s + \ln\big(1 + k_s\hat{x}_s(j)\big) + \sum_{u\in N(s)} \ln\big(1 + \hat{m}_{us}(j)\big)$$

We then use the approximation $\ln(1+\epsilon) \approx \epsilon$ for small $\epsilon$:

$$k\hat{y}_s(j) \leftarrow -\ln Z_s + k\hat{x}_s(j) + \sum_{u\in N(s)} \hat{m}_{us}(j) \tag{10}$$

Summing both sides over $j$ gives us:

$$k_s\underbrace{\sum_j \hat{y}_s(j)}_{=0} \leftarrow -k_s\ln Z_s + k_s\underbrace{\sum_j \hat{x}_s(j)}_{=0} + \underbrace{\sum_j\sum_{u\in N(s)}\hat{m}_{us}(j)}_{=0}$$

Hence, we see that $\ln Z_s$ needs to be 0, and therefore our normalizer is actually a normalization constant and for all nodes $Z_s = 1$. Plugging $Z_s = 1$ back into Eq. (10) leads to Eq. (8):

$$\hat{y}_s(j) \leftarrow \hat{x}_s(j) + \frac{1}{k_s}\sum_{u\in N(s)}\hat{m}_{us}(j)$$

($ii$) Equation (9): Using Lemma 10, we can write Eq. (2) as follows (recall that $k_t$ and $\psi'_{st}$ take care of the normalization):

$$m_{st}(i) \leftarrow k_t\sum_j \psi'_{st}(j,i)\,x_s(j)\prod_{u\in N(s)\setminus t} m_{us}(j)$$

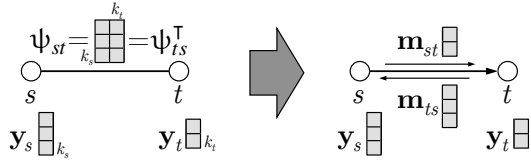

Figure 6: Matrix $\psi_{st}$ represents the edge potential from Eq. (15) and implies the direction $s \to t$. By ignoring the "echo cancellation" term in Eq. (16), one can think of the messages as $\mathbf{m}_{st} \propto \psi_{st}^\mathsf{T} \mathbf{y}_s$ and $\mathbf{m}_{ts} \propto \psi_{st} \mathbf{y}_t$.

By further using Eq. (8), we get:

$$\leftarrow k_t \sum_j \psi'_{st}(j,i) \frac{x_s(j) \prod_{u \in N(s)} m_{us}(j)}{m_{ts}(j)}$$

$$\leftarrow k_t \sum_j \psi'_{st}(j,i) \frac{y_s(j)}{m_{ts}(j)}$$

Then, using the centering together with the approximation $\frac{\frac{1}{k}+\epsilon_1}{1+\epsilon_2} \approx \frac{1}{k} + \epsilon_1 - \frac{1}{k}\epsilon_2$, we get:

$$1 + \hat{m}_{st}(i) \leftarrow k_t \sum_j \left(\frac{1}{k_t} + \hat{\psi}'_{st}(j,i)\right) \frac{\frac{1}{k_s} + \hat{y}_s(j)}{1 + \hat{m}_{ts}(j)}$$

$$\leftarrow k_t \sum_j \left(\frac{1}{k_t} + \hat{\psi}'_{st}(j,i)\right)\left(\frac{1}{k_s} + \hat{y}_s(j) - \frac{1}{k_s}\hat{m}_{ts}(j)\right)$$

$$\leftarrow k_t \Big(\frac{1}{k_t} + \underbrace{\frac{1}{k_s}\sum_j \hat{\psi}'_{st}(j,i)}_{=\hat{c}'_{st}(i)} + \underbrace{\frac{1}{k_t}\sum_j \hat{y}_s(j)}_{=0} + \sum_j \hat{\psi}'_{st}(j,i)\hat{y}_s(j)$$

$$- \underbrace{\frac{1}{k_s k_t}\sum_j \hat{m}_{ts}(j)}_{0} - \frac{1}{k_s}\sum_j \hat{\psi}'_{st}(j,i)\hat{m}_{ts}(j)\Big)$$

$$\hat{m}_{st}(i) \leftarrow \frac{k_t}{k_s}\hat{c}'_i + k_t \sum_j \hat{\psi}'_{st}(j,i)\hat{y}_s(j) - \frac{k_t}{k_s}\sum_j \hat{\psi}'_{st}(j,i)\hat{m}_{ts}(j) \qquad \square$$

Equations (8) and (9) can be written in matrix notation as:

$$\boxed{\hat{\mathbf{y}}_s \leftarrow \left(\hat{\mathbf{x}}_s + \frac{1}{k_s} \cdot \sum_{u \in N(s)} \hat{\mathbf{m}}_{us}\right)} \qquad (11)$$

$$\boxed{\hat{\mathbf{m}}_{st} \leftarrow \frac{k_t}{k_s}\hat{\mathbf{c}}'_{st} + k_t \hat{\psi}'^{\mathsf{T}}_{st}\left(\hat{\mathbf{y}}_s - \frac{1}{k_s}\hat{\mathbf{m}}_{ts}\right)} \qquad (12)$$

An alternative way to write the message updates is

$$\hat{\mathbf{m}}_{st} \leftarrow \frac{k_t}{k_s}\hat{\mathbf{c}}'_{st} + k_t \hat{\psi}'^{\mathsf{T}}_{st}\left(\hat{\mathbf{x}}_s + \frac{1}{k_s} \cdot \sum_{u \in N(s)\setminus t} \hat{\mathbf{m}}_{us}\right) \qquad (13)$$

It is instructive to compare the above derived linearized update equations against the matrix formulations of the original BP update equations Eq. (1) and Eq. (2): by using the symbol $\odot$ for the Hadamard product[10], those can be written compactly in matrix notation, as:

$$\boxed{\mathbf{y}_s \leftarrow \frac{1}{Z_s}\left(\mathbf{x}_s \odot \bigodot_{u \in N(s)} \mathbf{m}_{us}\right)} \qquad (14)$$

$$\mathbf{m}_{st} \leftarrow \frac{1}{Z_{st}}\psi^{\mathsf{T}}_{st}\left(\mathbf{x}_s \odot \bigodot_{u \in N(s)\setminus t} \mathbf{m}_{us}\right) \qquad (15)$$

---
[10]The Hadamard product is defined by: $\mathbf{Z} = \mathbf{X} \odot \mathbf{Y} \Leftrightarrow Z(i,j) = X(i,j) \cdot Y(i,j)$.

Notice that the potential $\boldsymbol{\psi}_{st}$ is represented by a $k_s \times k_t$-dimensional *"compatibility matrix"* and that the transpose $\boldsymbol{\psi}_{st}^\mathsf{T} = \boldsymbol{\psi}_{ts}$ (see Fig. 6). This follows from the definition of a potential in a pairwise MRF and the resulting derivation of belief propagation (Yedidia, Freeman, and Weiss 2003). Also notice that we could reduce the amount of necessary calculation by first multiplying all incoming messages at a node, and then dividing through the message that a node sends to itself via a neighbor (we call this compensation *"echo cancellation"*). This approach is also called *"message-passing with division"* (Koller and Friedman 2009) (or "belief-update message passing") and can be made precise by defining a component-wise division operator by: $\mathbf{Z} = \mathbf{X} \oslash \mathbf{Y} \Leftrightarrow Z(i,j) = X(i,j)/Y(i,j)$ where $0/0 = 0$. Equation (15) can then be written more concisely as:

$$\boxed{\mathbf{m}_{st} \leftarrow \frac{1}{Z_{st}} \boldsymbol{\psi}_{st}^\mathsf{T} \big(\mathbf{y}_s \oslash \mathbf{m}_{ts}\big)} \tag{16}$$

We invite the reader to carefully compare Eqs. (11) and (12) with the original BP update equations Eqs. (11) and (13). Notice that the first term $\frac{k_t}{k_s}\hat{\mathbf{c}}'_{st}$ in Eq. (12) vanishes in the case of doubly stochastic potentials (or more generally, potentials with equal column and row sums). For non-doubly stochastic potentials, this term captures the prior probabilities of node classes resulting from non-equivalent row or column sums.

**Steady state messages**

From Lemma 12, we can derive a closed-form equation for the message in steady-state of belief propagation.

**Lemma 13** (Steady state messages). *After convergence of belief propagation, message propagation can be approximated in terms of the steady centered beliefs as:*

$$\boxed{\hat{\mathbf{m}}_{st} = \frac{k_t}{k_s}\hat{\mathbf{c}}'_{st} + k_t \hat{\boldsymbol{\psi}}_{st}^{'\mathsf{T}}\big(\hat{\mathbf{y}}_s - \hat{\boldsymbol{\psi}}_{st}^{''}\hat{\mathbf{y}}_t\big)} \tag{17}$$

*Proof Lemma 13.* To increase the readability of this proof, we ignore the subscripts in $\boldsymbol{\psi}_{st}$, $\mathbf{c}_{st}$, $\mathbf{r}_{st}$, and write instead $\boldsymbol{\psi}$, $\mathbf{c}$, $\mathbf{r}$, respectively. We start by writing Eq. (9) for the messages in each of the two directions across the same edge $s-t$:

$$\hat{m}_{st}(i) \leftarrow \frac{k_t}{k_s}\hat{c}(i)' + k_t \sum_{j=1}^{k_s} \hat{\psi}'(j,i)\Big(\hat{y}_s(j) - \frac{1}{k_s}\hat{m}_{ts}(j)\Big)$$

$$\hat{m}_{ts}(j) \leftarrow \frac{k_s}{k_t}\hat{r}(j)'' + k_s \sum_{g=1}^{k_t} \hat{\psi}''(g,j)\Big(\hat{y}_t(g) - \frac{1}{k_t}\hat{m}_{st}(g)\Big)$$

We then simply combine both equations into one:

$$\hat{m}_{st}(i) \leftarrow \frac{k_t}{k_s}\hat{c}(i)' + k_t \sum_{j=1}^{k_s} \hat{\psi}'(j,i)\,\hat{y}_s(j) - \frac{k_t}{k_s}\sum_{j=1}^{k_s}\hat{\psi}'(j,i) \cdot$$

$$\Big(\frac{k_s}{k_t}\hat{r}(j)'' + k_s\sum_{g=1}^{k_t}\hat{\psi}''(g,j)\,\hat{y}_t(g) - \frac{k_s}{k_t}\sum_{g=1}^{k_t}\hat{\psi}''(g,j)\,\hat{m}_{st}(g)\Big)$$

Now, if the equations converge, then $\hat{m}_{st}(g)$ on both the left and right side of the equation need to be equivalent. We can, therefore, replace the update symbol with equality and group related terms together:

$$\hat{m}_{st}(i) - \frac{k_t \cancel{k_s}}{\cancel{k_s} k_t}\sum_{j=1}^{k_s}\hat{\psi}'(j,i)\sum_{g=1}^{k_t}\hat{\psi}''(g,j)\,\hat{m}_{st}(g) =$$

$$\frac{k_t}{k_s}\hat{c}(i)' - \frac{k_t \cancel{k_s}}{\cancel{k_s} k_t}\sum_{j=1}^{k_s}\hat{\psi}'(j,i)\,\hat{r}(j)'' + k_t\sum_{j=1}^{k_s}\hat{\psi}'(j,i)\,\hat{y}_s(j) - \frac{k_t \cancel{k_s}}{\cancel{k_s}}\sum_{j=1}^{k_s}\hat{\psi}'(j,i)\sum_{g=1}^{k_t}\hat{\psi}''(g,j)\,\hat{y}_t(g)$$

With $\mathbf{I}_{k_t}$ as the $k_t$-dimensional identity matrix, this can then be written in matrix notation as:

$$(\mathbf{I}_{k_t} - \hat{\boldsymbol{\psi}}^{'\mathsf{T}}\hat{\boldsymbol{\psi}}'')\hat{\mathbf{m}}_{st} = \frac{k_t}{k_s}\hat{\mathbf{c}}' - \hat{\boldsymbol{\psi}}^{'\mathsf{T}}\hat{\mathbf{r}}'' + k_t\hat{\boldsymbol{\psi}}^{'\mathsf{T}}\hat{\mathbf{y}}_s - k_t\hat{\boldsymbol{\psi}}^{'\mathsf{T}}\hat{\boldsymbol{\psi}}''\hat{\mathbf{y}}_t$$

Recall that $\hat{\mathbf{c}}' = \hat{\boldsymbol{\psi}}'^{\mathsf{T}} \mathbf{1}_{k_s}$ and $\hat{\mathbf{r}}'' = \hat{\boldsymbol{\psi}}'' \mathbf{1}_{k_t}$.

$$(\mathbf{I}_{k_t} - \hat{\boldsymbol{\psi}}'^{\mathsf{T}} \hat{\boldsymbol{\psi}}'') \hat{\mathbf{m}}_{st} = \big(\frac{k_t}{k_s} \hat{\boldsymbol{\psi}}'^{\mathsf{T}} \mathbf{1}_{k_s} - \hat{\boldsymbol{\psi}}'^{\mathsf{T}} \hat{\boldsymbol{\psi}}'' \mathbf{1}_{k_t}\big) + k_t \hat{\boldsymbol{\psi}}'^{\mathsf{T}} \hat{\mathbf{y}}_s - k_t \hat{\boldsymbol{\psi}}'^{\mathsf{T}} \hat{\boldsymbol{\psi}}'' \hat{\mathbf{y}}_t$$

$$= \hat{\boldsymbol{\psi}}'^{\mathsf{T}} \big(\frac{k_t}{k_s} \mathbf{1}_{k_s} + k_t \hat{\mathbf{y}}_s\big) - \hat{\boldsymbol{\psi}}'^{\mathsf{T}} \hat{\boldsymbol{\psi}}'' \big(\mathbf{1}_{k_t} + k_t \hat{\mathbf{y}}_t\big)$$

If all entries of $\hat{\boldsymbol{\psi}}$ are appropriately small (so that the spectral radius $\rho(\hat{\boldsymbol{\psi}}'^{\mathsf{T}} \hat{\boldsymbol{\psi}}'') < 1$), then the inverse of $(\mathbf{I}_{k_t} - \hat{\boldsymbol{\psi}}'^{\mathsf{T}} \hat{\boldsymbol{\psi}}'')$ exists. Thus, by further substituting $\hat{\boldsymbol{\psi}}'^{\mathsf{T}}_{\triangle} := (\mathbf{I}_{k_t} - \hat{\boldsymbol{\psi}}'^{\mathsf{T}} \hat{\boldsymbol{\psi}}'')^{-1} \hat{\boldsymbol{\psi}}'^{\mathsf{T}}$, we can write:

$$\hat{\mathbf{m}}_{st} = \hat{\boldsymbol{\psi}}'^{\mathsf{T}}_{\triangle} \big(\frac{k_t}{k_s} \mathbf{1}_{k_s} + k_t \hat{\mathbf{y}}_s\big) - \hat{\boldsymbol{\psi}}'^{\mathsf{T}}_{\triangle} \hat{\boldsymbol{\psi}}'' \big(\mathbf{1}_{k_t} + k_t \hat{\mathbf{y}}_t\big) \tag{18}$$

$$= \hat{\boldsymbol{\psi}}'^{\mathsf{T}}_{\triangle} \big(\frac{k_t}{k_s} \mathbf{1}_{k_s} - \hat{\boldsymbol{\psi}}'' \mathbf{1}_{k_t}\big) + k_t \hat{\boldsymbol{\psi}}'^{\mathsf{T}}_{\triangle} \hat{\mathbf{y}}_s - k_t \hat{\boldsymbol{\psi}}'^{\mathsf{T}}_{\triangle} \hat{\boldsymbol{\psi}}'' \hat{\mathbf{y}}_t$$

By further substituting $\hat{\mathbf{h}}' := \frac{k_t}{k_s} \hat{\boldsymbol{\psi}}'^{\mathsf{T}}_{\triangle} \mathbf{1}_{k_s} - \hat{\boldsymbol{\psi}}'^{\mathsf{T}}_{\triangle} \hat{\boldsymbol{\psi}}'' \mathbf{1}_{k_t}$, we get the following equation for the message updates Eq. (12) after convergence:

$$\boxed{\hat{\mathbf{m}}_{st} = \hat{\mathbf{h}}' + k_t \hat{\boldsymbol{\psi}}'^{\mathsf{T}}_{\triangle} (\hat{\mathbf{y}}_s - \hat{\boldsymbol{\psi}}'' \hat{\mathbf{y}}_t)} \tag{19}$$

Next notice that $\boldsymbol{\psi}'^{\mathsf{T}}_{\triangle} \approx \boldsymbol{\psi}'^{\mathsf{T}}$, since $\mathbf{I}_{k_t} \gg |\hat{\boldsymbol{\psi}}'^{\mathsf{T}} \hat{\boldsymbol{\psi}}''|$, and therefore $(\mathbf{I}_{k_t} - \hat{\boldsymbol{\psi}}'^{\mathsf{T}} \hat{\boldsymbol{\psi}}'')^{-1} \approx \mathbf{I}_{k_t}$. From this, we can now also approximate $\hat{\mathbf{h}}' \approx \frac{k_t}{k_s} (\hat{\boldsymbol{\psi}}'^{\mathsf{T}} \mathbf{1}_{k_s}) - \hat{\boldsymbol{\psi}}'^{\mathsf{T}} (\hat{\boldsymbol{\psi}}'' \mathbf{1}_{k_t}) = \frac{k_t}{k_s} \hat{\mathbf{c}}' - \hat{\boldsymbol{\psi}}'^{\mathsf{T}} \hat{\mathbf{r}}''$. Further ignoring the second term, we get $\hat{\mathbf{h}}' \approx \frac{k_t}{k_s} \hat{\mathbf{c}}'$. Plugging back into Eq. (19) finally gives us Eq. (17). $\square$

Also notice that we can alternatively write Eq. (18) as function of the uncentered beliefs, which results in a very intuitive equation:

$$\hat{\mathbf{m}}_{st} = k_t \hat{\boldsymbol{\psi}}'^{\mathsf{T}}_{\triangle} \big(\frac{1}{k_s} \mathbf{1}_{k_s} + \hat{\mathbf{y}}_s\big) - k_t \hat{\boldsymbol{\psi}}'^{\mathsf{T}}_{\triangle} \hat{\boldsymbol{\psi}}'' \big(\frac{1}{k_t} \mathbf{1}_{k_t} + \hat{\mathbf{y}}_t\big)$$

$$= k_t \hat{\boldsymbol{\psi}}'^{\mathsf{T}}_{\triangle} (\mathbf{y}_s) - k_t \hat{\boldsymbol{\psi}}'^{\mathsf{T}}_{\triangle} \hat{\boldsymbol{\psi}}'' (\mathbf{y}_t)$$

$$= k_t \hat{\boldsymbol{\psi}}'^{\mathsf{T}}_{\triangle} (\mathbf{y}_s - \hat{\boldsymbol{\psi}}'' \mathbf{y}_t)$$

**Theorem 4: The actual linearization**

Finally, by using matrix notation, we can transform and write Eq. (17) for all nodes and edges together as one large equation system and get Theorem 4.

*Proof Theorem 4.* For steady-state, we can write Eq. (8) in vector form as:

$$\hat{\mathbf{y}}_s = \hat{\mathbf{x}}_s + \frac{1}{k_s} \sum_{u \in N(s)} \hat{\mathbf{m}}_{us}$$

By permuting subscripts, we can also write Eq. (17) as

$$\hat{\mathbf{m}}_{us} = \frac{k_s}{k_u} \hat{\mathbf{c}}'_{us} + k_s \hat{\boldsymbol{\psi}}^{\mathsf{T}}_{us} (\hat{\mathbf{y}}_u - \hat{\boldsymbol{\psi}}''_{us} \hat{\mathbf{y}}_s)$$

Combining the last two equations, we get

$$\hat{\mathbf{y}}_s = \underbrace{\hat{\mathbf{x}}_s}_{\text{1st}} + \underbrace{\sum_{u \in N(s)} \frac{\hat{\mathbf{c}}'_{us}}{k_u}}_{\text{2nd}} + \underbrace{\sum_{u \in N(s)} \hat{\boldsymbol{\psi}}'^{\mathsf{T}}_{us} \hat{\mathbf{y}}_u}_{\text{3rd}} - \underbrace{\sum_{u \in N(s)} \hat{\boldsymbol{\psi}}'^{\mathsf{T}}_{us} \hat{\boldsymbol{\psi}}''_{us} \hat{\mathbf{y}}_s}_{\text{4th}} \tag{20}$$

By using our combined new vectors and matrices $\hat{\mathbf{y}}$, $\hat{\mathbf{x}}$, $\mathbf{k}$, and $\hat{\boldsymbol{\psi}}$ (and analogously for the column-recentered residual matrix $\hat{\boldsymbol{\psi}}''$), we can write Eq. (20) in matrix form as:

$$\hat{\mathbf{y}} = \hat{\mathbf{X}} + \hat{\boldsymbol{\psi}}'^{\mathsf{T}} \mathbf{k} + \hat{\boldsymbol{\psi}}'^{\mathsf{T}} \hat{\mathbf{y}} - \hat{\boldsymbol{\psi}}'^{\mathsf{T}} \hat{\boldsymbol{\psi}}'' \hat{\mathbf{y}}$$

From Corollary 8, we know $\hat{\boldsymbol{\psi}}'^{\mathsf{T}}_{ij} = \hat{\boldsymbol{\psi}}''_{ji}$. Therefore, from our construction we also have $\hat{\boldsymbol{\psi}}'^{\mathsf{T}} = \hat{\boldsymbol{\psi}}''$. We thus get

$$\hat{\mathbf{y}} = \hat{\mathbf{x}} + \hat{\boldsymbol{\psi}}'^{\mathsf{T}}\mathbf{k} + \hat{\boldsymbol{\psi}}'^{\mathsf{T}}\hat{\mathbf{y}} - \hat{\boldsymbol{\psi}}'^{\mathsf{T}2}\hat{\mathbf{y}} \tag{21}$$

Equation (21) is now a straight-forward linear equation system that can be solved for $\hat{\mathbf{y}}$.

To finish the proof, first notice that each of our approximations become exact for $\lim_{\epsilon \to 0^+}$. Second, notice that higher order deltas vanish and our equation simplify to $\hat{\mathbf{y}} = \hat{\mathbf{x}} + \hat{\boldsymbol{\psi}}'^{\mathsf{T}}\mathbf{k} + \hat{\boldsymbol{\psi}}'^{\mathsf{T}}\hat{\mathbf{y}}$. While the individual beliefs go to zero during this limit consideration, their relative sizes stay the same, and thus the Maximum Marginal for each node stays the same. □

**Proposition 5: Update equations and convergence**

*Proof Proposition 5.* From the Jacobi method for solving linear systems (Saad 2003) – also known as the Neumann series expansion of the matrix inverse – we know that the solution for $\mathbf{y} = (\mathbf{I} - \mathbf{M})^{-1}\mathbf{x}$ can be calculated (under certain conditions) via the iterative update equation

$$\mathbf{y}^{(r+1)} \leftarrow \mathbf{x} + \mathbf{M}\,\mathbf{y}^{(r)}$$

These updates are known to converge for any choice of initial values for $\mathbf{y}^{(0)}$, as long as $\mathbf{M}$ has a spectral radius $\rho(\mathbf{M}) < 1$. The same convergence guarantees carry over to Eq. (5). We thus know that the update equation Eq. (5) converges if and only if the spectral radius of the matrix $\hat{\boldsymbol{\psi}}'^{\mathsf{T}} - \hat{\boldsymbol{\psi}}'^{\mathsf{T}2}$ is smaller than 1. □

## C  Special formulations

In this section, we derive alternative formulations of Theorem 4 for increasingly specialized cases: Graphs with node and edge types (Appendix C), graphs with equal number of classes for all nodes (Appendix C), graphs with one single directed edge potential (Appendix C), and the special case described in prior work of one single symmetric, doubly stochastic potential (Appendix C).

**Node types and edge types with repeated potentials**

In many realistic scenarios, the number of edges is usually larger than the number of different edge types (or edge potentials). For example, assume a set $Q$ of different node types.[11] We then have a $|Q|$-partite network and each node with type $q \in Q$ can be one of $k(q)$ classes. Further assume that the couplings along an edge only depend on the types at both ends of the edge. Then there are maximal $|Q|^2$ different row-recentered potentials irrespective of the size of the network, whereas the most general formulation of Theorem 4 would redundantly store in $\hat{\boldsymbol{\psi}}'$ one full potential for each edge (Recall that we have one row-recentered potential for every edge direction, thus two for every edge type). In the following, we transform the update equations so that every different row-recentered edge potential appears only once in the equations. Notice that the ensuing formulation allows for more than one potential between any pair of node types.

A complication in deriving a compact matrix formulation is that different types of nodes may have different numbers of classes. We address this issue by creating separate matrices that contain the beliefs of nodes with the same number of classes. Concretely, let $N = \{1, 2, \ldots, n\}$ be the set of all nodes, $q(s)$ be the type of node $s$, $k(q)$ be the number of classes for type $q$, and $K = \{k(1), k(2), \ldots, k(|Q|)\}$ be the set of *numbers* of classes across all nodes.[12] Let $N_k \subseteq N$ denote the set of nodes with $k \in K$ classes so that all nodes are partitioned into groups $N_{k_1}, N_{k_2}, \ldots, N_{k_{|K|}}$. Let $n_k = |N_k|$ denote the number of nodes with $k$ classes. We assume a numbering of nodes such that $N_{k_1} = \{1, 2, \ldots, n_{k_1}\}$, $N_{k_2} = \{n_{k_1} + 1, n_{k_1} + 2, \ldots, n_{k_1} + n_{k_2}\}$, and so on. Given this convention, each node $s$ has a unique order $o(s)$ within the set $N_{k(q(s))}$. For each $k \in K$, we create two $n_k \times k$ matrices $\hat{\mathbf{Y}}_k$ and $\hat{\mathbf{X}}_k$ that contain the posterior and prior residual beliefs of all nodes with $k$ classes.

For each potential $\boldsymbol{\psi} \in \mathbb{R}^{\ell \times k}$, we create two centered residual potentials $\hat{\boldsymbol{\psi}}' \in \mathbb{R}^{\ell \times k}$ and $\hat{\boldsymbol{\psi}}''^{\mathsf{T}} = (\hat{\boldsymbol{\psi}}^{\mathsf{T}})' \in \mathbb{R}^{k \times \ell}$ that correspond to the two modulations across the two directions of an edge. For notational convenience, we treat them as two distinct potentials and ignore their common ancestry. For example, $\boldsymbol{\psi}_{12} \in \mathbb{R}^{3 \times 2}$ leads to $\hat{\boldsymbol{\psi}}'_{12} \in \mathbb{R}^{3 \times 2}$ and $\hat{\boldsymbol{\psi}}'_{21} = \hat{\boldsymbol{\psi}}''^{\mathsf{T}}_{12} \in \mathbb{R}^{2 \times 3}$. For each newly created row-recentered residual potential $\hat{\boldsymbol{\psi}}' \in \mathbb{R}^{\ell \times k}$, we create two new matrices: (*i*) the adjacency matrix $\mathbf{W}_{\hat{\boldsymbol{\psi}}'} \in \mathbb{R}^{n_\ell \times n_k}$ with $W_{\hat{\boldsymbol{\psi}}'}(o(s), o(t)) = 1$ if node $s$ with $\ell$ classes is connected to node $t$ with $k$ classes via an edge potential $\hat{\boldsymbol{\psi}}'$; and (*ii*) the diagonal in-degree matrix $\mathbf{D}^{\text{in}}_{\hat{\boldsymbol{\psi}}'} \in \mathbb{R}^{n_k \times n_k}$ with $D^{\text{in}}_{\hat{\boldsymbol{\psi}}'}(o(t), o(t)) = d$ if there are $d$ different nodes $s$ that are connected to $t$ via an edge potential $\hat{\boldsymbol{\psi}}'$ (notice that $\hat{\boldsymbol{\psi}}'$ modulates along the direction $s \to t$, therefore "in-degree" at node $t$ with $k$ classes).

---

[11] Notice our use of vocabulary: the "type" of a node in a network is observed and known a priori (e.g., whether the node represents a user or a product), whereas the "class" of a node is the label that we are trying to learn (e.g., whether the user is male or female).

[12] In a slight abuse of set notation, we allow here e.g. $\{1, 1, 2\}$ to stand for a set $\{1, 2\}$.

**Proposition 14** (Edge types). *Let $\hat{\mathcal{P}}'$ bet the set of all row-recentered potentials, $\hat{\mathcal{P}}'^{\ell \times k} \subseteq \mathcal{P}'$ be the subset with dimensions $\ell \times k$, and $\hat{\mathbf{Y}}_k$, $\hat{\mathbf{X}}_k$, $\mathbf{W}_{\hat{\psi}'}$, $\mathbf{D}_{\hat{\psi}'}^{\text{in}}$ be the above defined partitioned matrices for all $k \in K$. For each $\hat{\psi}' \in \hat{\mathcal{P}}'$ let $\hat{\psi}''$ be the corresponding column-recentered potential. The update equation Eq. (5) can then be written as follows:*

$$\forall k \in K : \hat{\mathbf{Y}}_k \leftarrow \hat{\mathbf{X}}_k + \hat{\mathbf{C}}'_{k*} + \sum_{\hat{\psi}' \in \hat{\mathcal{P}}'^{\ell \times k}, \ell \in K} \left( \mathbf{W}_{\hat{\psi}'}^\intercal \hat{\mathbf{Y}}_\ell \hat{\psi}' - \mathbf{D}_{\hat{\psi}'}^{\text{in}} \hat{\mathbf{Y}}_k \hat{\psi}''^\intercal \hat{\psi}' \right) \quad (22)$$

*with*

$$\hat{\mathbf{C}}'_{k*} := \sum_{\hat{\psi}' \in \hat{\mathcal{P}}'^{\ell \times k}, \ell \in K} \frac{1}{\ell} \mathbf{W}_{\hat{\psi}'}^\intercal \mathbf{1}_{n_\ell} \mathbf{1}_\ell^\intercal \hat{\psi}'$$

*Proof Proposition 14.* We are going to derive Eq. (22) from Eq. (20). For convenience, we repeat here both equations:

$$\hat{\mathbf{Y}}_k = \hat{\mathbf{X}}_k + \hat{\mathbf{C}}'_{k*} + \sum_{\hat{\psi}' \in \hat{\mathcal{P}}'^{\ell \times k}, \ell \in K} \left( \mathbf{W}_{\hat{\psi}'}^\intercal \hat{\mathbf{Y}}_\ell \hat{\psi}' - \mathbf{D}_{\hat{\psi}'}^{\text{in}} \hat{\mathbf{Y}}_k \hat{\psi}''^\intercal \hat{\psi}' \right)$$

$$\hat{\mathbf{y}}_s = \underbrace{\hat{\mathbf{x}}_s}_{\text{1st}} + \underbrace{\sum_{u \in N(s)} \frac{\hat{\mathbf{c}}'_{us}}{k_u}}_{\text{2nd}} + \underbrace{\sum_{u \in N(s)} \hat{\psi}'^\intercal_{us} \hat{\mathbf{y}}_u}_{\text{3rd}} - \underbrace{\sum_{u \in N(s)} \hat{\psi}'^\intercal_{us} \hat{\psi}''_{us} \hat{\mathbf{y}}_s}_{\text{4th}}$$

We need to show that any vector $\hat{\mathbf{y}}_s^\intercal$ for a node $s$ with $k$ classes is equivalent to the $o(s)$-th row of $\hat{\mathbf{Y}}_k$, for which we are going to write $\hat{\mathbf{Y}}_k(o(s), :)$ from now on (recall that $o(s)$ is the order of node $s$ within $N_k$). We show the equivalence for each the 4 terms separately:

1st term: $\hat{\mathbf{x}}_s^\intercal = \hat{\mathbf{X}}_k(o(s), :)$ by construction.

2nd term: For the following, recall that $\hat{\mathbf{c}}'_{us} = \hat{\psi}'^\intercal_{us} \mathbf{1}_{k_u}$:

$$\left( \sum_{u \in N(s)} \frac{1}{k_u} \hat{\mathbf{c}}'_{us} \right)^\intercal = \sum_{u \in N(s)} \frac{1}{k_u} \hat{\mathbf{c}}'^\intercal_{us}$$

$$= \sum_{u \in N(s)} \frac{1}{k_u} \mathbf{1}_{k_u}^\intercal \hat{\psi}'_{us}$$

$$= \sum_{u \in N(s)} \frac{1}{k_u} W_{\hat{\psi}'_{us}}^\intercal (o(s), o(u)) \mathbf{1}_{k_u}^\intercal \hat{\psi}'_{us}$$

$$= \sum_{\hat{\psi}' \in \hat{\mathcal{P}}'^{\ell \times k}, \ell \in K} \frac{1}{\ell} \mathbf{W}_{\hat{\psi}'}^\intercal (o(s), :) \mathbf{1}_{n_\ell} \mathbf{1}_\ell^\intercal \hat{\psi}'$$

$$= \hat{\mathbf{C}}'_{k*}(o(s), :)$$

3rd term:

$$\left( \sum_{u \in N(s)} \hat{\psi}'^\intercal_{us} \hat{\mathbf{y}}_u \right)^\intercal = \sum_{u \in N(s)} \hat{\mathbf{y}}_u^\intercal \hat{\psi}'_{us}$$

$$= \sum_{u \in N(s)} \hat{\mathbf{Y}}_{k_u}(o(u), :) \hat{\psi}'_{us}$$

$$= \sum_{u \in N(s)} W_{\hat{\psi}'_{us}}^\intercal (o(s), o(u)) \hat{\mathbf{Y}}_{k_u}(o(u), :) \hat{\psi}'_{us}$$

$$= \sum_{\hat{\psi}' \in \hat{\mathcal{P}}'^{\ell \times k}, \ell \in K} \mathbf{W}_{\hat{\psi}'}^\intercal (o(s), :) \hat{\mathbf{Y}}_\ell \hat{\psi}'$$

$$= \left( \sum_{\hat{\psi}' \in \hat{\mathcal{P}}'^{\ell \times k}, \ell \in K} \mathbf{W}_{\hat{\psi}'}^\intercal \hat{\mathbf{Y}}_\ell \hat{\psi}' \right)(o(s), :)$$

4th term:

$$\left(\sum_{u\in N(s)} \hat{\boldsymbol{\psi}}'^{\mathsf{T}}_{us}\hat{\boldsymbol{\psi}}''_{us}\hat{\mathbf{y}}_s\right)^{\mathsf{T}} = \sum_{u\in N(s)} \hat{\mathbf{y}}^{\mathsf{T}}_s\hat{\boldsymbol{\psi}}''^{\mathsf{T}}_{us}\hat{\boldsymbol{\psi}}'_{us}$$

$$= \sum_{u\in N(s)} \hat{\mathbf{Y}}_k(o(s),:)\hat{\boldsymbol{\psi}}''^{\mathsf{T}}_{us}\hat{\boldsymbol{\psi}}'_{us}$$

$$= \sum_{\hat{\boldsymbol{\psi}}'\in\hat{\mathcal{P}}'^{\ell\times k},\ell\in K} D^{\text{in}}_{\hat{\boldsymbol{\psi}}'}(o(s),o(s))\,\hat{\mathbf{Y}}_k(o(s),:)\hat{\boldsymbol{\psi}}''^{\mathsf{T}}\hat{\boldsymbol{\psi}}'$$

$$= \left(\sum_{\hat{\boldsymbol{\psi}}'\in\hat{\mathcal{P}}'^{\ell\times k},\ell\in K} \mathbf{D}^{\text{in}}_{\hat{\boldsymbol{\psi}}'}\hat{\mathbf{Y}}_k\hat{\boldsymbol{\psi}}''^{\mathsf{T}}\hat{\boldsymbol{\psi}}'\right)(o(s),:) \qquad \square$$

### Nodes with same number of classes $k$

Proposition 14 simplifies considerably when all nodes have the same number of classes $k$:

**Corollary 15** (Same $k$). *Let $k$ be the number of classes for each node in the graph, $\hat{\mathcal{P}}'$ the set of row-recentered residual edge potentials (all with $k\times k$ dimensions), $\hat{\mathbf{Y}}$ and $\hat{\mathbf{X}}$ the $n\times k$ dimensional final and explicit belief matrices, and $\mathbf{W}_{\hat{\boldsymbol{\psi}}'}$ and $\mathbf{D}^{\text{in}}_{\hat{\boldsymbol{\psi}}'}$ the adjacency and in-degree matrices for each potential $\hat{\boldsymbol{\psi}}\in\hat{\mathcal{P}}'$. The update equations can then be simplified to:*

$$\hat{\mathbf{Y}}\leftarrow \hat{\mathbf{X}} + \hat{\mathbf{C}}'_* + \sum_{\hat{\boldsymbol{\psi}}'\in\hat{\mathcal{P}}'}\left(\mathbf{W}^{\mathsf{T}}_{\hat{\boldsymbol{\psi}}'}\hat{\mathbf{Y}}\hat{\boldsymbol{\psi}}' - \mathbf{D}^{\text{in}}_{\hat{\boldsymbol{\psi}}'}\hat{\mathbf{Y}}\hat{\boldsymbol{\psi}}''^{\mathsf{T}}\hat{\boldsymbol{\psi}}'\right) \qquad (23)$$

*with*

$$\hat{\mathbf{C}}'_* := \frac{1}{k}\sum_{\hat{\boldsymbol{\psi}}'\in\hat{\mathcal{P}}'} \mathbf{W}^{\mathsf{T}}_{\hat{\boldsymbol{\psi}}'}\mathbf{1}_n\mathbf{1}^{\mathsf{T}}_k\hat{\boldsymbol{\psi}}'$$

Also the convergence criterium and the closed-form solution allow very concise formulations. For this step we need to introduce two new notations: Let $\mathbf{x}_j$ denote the $j$-th column of matrix $\mathbf{X}$ (i.e., $\mathbf{X} = \{x_{ij}\} = [\mathbf{x}_1\ldots\mathbf{x}_n]$) and let $\mathbf{X}$ and $\mathbf{Y}$ be matrices of order $m\times n$ and $p\times q$, respectively. First, the *vectorization* of a matrix $\mathbf{X}$ stacks its columns one underneath the other to form a single column vector:

$$\texttt{vec}\,(\mathbf{X}) = \begin{bmatrix}\mathbf{x}_1\\ \vdots\\ \mathbf{x}_n\end{bmatrix}$$

Second, the *Kronecker product* ($\otimes$) of $\mathbf{X}$ and $\mathbf{Y}$ results in a $mp\times nq$ block matrix:

$$\mathbf{X}\otimes\mathbf{Y} = \begin{bmatrix} x_{11}\mathbf{Y} & x_{12}\mathbf{Y} & \ldots & x_{1n}\mathbf{Y}\\ \vdots & \vdots & \ddots & \vdots\\ x_{m1}\mathbf{Y} & x_{m2}\mathbf{Y} & \ldots & x_{mn}\mathbf{Y}\end{bmatrix}$$

With these notations, Corollary 6 now becomes

**Proposition 16** (Convergence with same $k$). *Update Eq. (23) converges if and only if the spectral radius $\rho(\mathbf{M}) < 1$ for*

$$\mathbf{M} := \sum_{\hat{\boldsymbol{\psi}}'\in\hat{\mathcal{P}}'}\left(\mathbf{W}^{\mathsf{T}}_{\hat{\boldsymbol{\psi}}'}\otimes\hat{\boldsymbol{\psi}}'^{\mathsf{T}} - \mathbf{D}^{\text{in}}_{\hat{\boldsymbol{\psi}}'}\otimes(\hat{\boldsymbol{\psi}}'^{\mathsf{T}}\hat{\boldsymbol{\psi}}'')\right)$$

*Furthermore, let $\hat{\mathbf{y}} := \texttt{vec}\big(\hat{\mathbf{Y}}^{\mathsf{T}}\big)$, $\hat{\boldsymbol{x}} := \texttt{vec}\big(\hat{\mathbf{X}}^{\mathsf{T}}\big)$, and $\hat{\mathbf{c}}'_* := \texttt{vec}\big(\hat{\mathbf{C}}'^{\mathsf{T}}_*\big)$. The closed-form solution of Eq. (23) is given by:*

$$\hat{\mathbf{y}} = (\mathbf{I}_{nk} - \mathbf{M})^{-1}(\hat{\boldsymbol{x}} + \hat{\mathbf{c}}'_*) \qquad (24)$$

Notice that Eq. (24) is a special case of Eq. (4) that factors out repeated edge potentials. This concise factorization with the Kronecker product is only possible if all nodes have the same number of classes $k$.

*Proof Proposition 16.* If all nodes have the same number of classes $k$ then all final and explicit beliefs form single $n\times k$ matrices

$\hat{\mathbf{Y}}$ and $\hat{\mathbf{X}}$. Furthermore, all potentials have $k \times k$ dimensions. Hence, Eq. (23) can be written as a single matrix equation:

$$\hat{\mathbf{Y}} = \hat{\mathbf{X}} + \hat{\mathbf{C}}_* + \sum_{\hat{\psi}' \in \mathcal{P}'} \left( \mathbf{W}_{\hat{\psi}'}^\mathsf{T} \hat{\mathbf{Y}} \hat{\psi}' - \mathbf{D}_{\hat{\psi}'} \hat{\mathbf{Y}} \hat{\psi}''^\mathsf{T} \hat{\psi}' \right)$$

$$\hat{\mathbf{Y}}^\mathsf{T} = \hat{\mathbf{X}}^\mathsf{T} + \hat{\mathbf{C}}_*^\mathsf{T} + \sum_{\hat{\psi}' \in \mathcal{P}'} \left( \hat{\psi}'^\mathsf{T} \hat{\mathbf{Y}}^\mathsf{T} \mathbf{W}_{\hat{\psi}'} - \hat{\psi}'^\mathsf{T} \hat{\psi}'' \hat{\mathbf{Y}}^\mathsf{T} \mathbf{D}_{\hat{\psi}'} \right)$$

with $\hat{\mathbf{C}}_*^\mathsf{T} := \frac{1}{k} \sum_{\hat{\psi}' \in \mathcal{P}'} \hat{\psi}'^\mathsf{T} \mathbf{W}_{\hat{\psi}'}$. We used the transpose in order for the later vectorization $\texttt{vec}$ to create vectors where the different beliefs of a node are adjacent (otherwise $\texttt{vec}(\hat{\mathbf{Y}})$ results in a vector where all beliefs in the same class from different nodes are adjacent). We next use *Roth's column lemma* (Henderson and Searle 1981; Roth 1934) that states that

$$\texttt{vec}(\mathbf{XYZ}) = (\mathbf{Z}^\mathsf{T} \otimes \mathbf{X}) \texttt{vec}(\mathbf{Y})$$

to rewrite this equation as

$$\hat{\mathbf{y}} = \hat{\mathbf{x}} + \hat{\mathbf{c}}'_* + \left( \sum_{\hat{\psi}' \in \mathcal{P}'} \left( \mathbf{W}_{\hat{\psi}'}^\mathsf{T} \otimes \hat{\psi}'^\mathsf{T} - \mathbf{D}_{\hat{\psi}'} \otimes (\hat{\psi}'^\mathsf{T} \hat{\psi}'') \right) \right) \hat{\mathbf{y}}$$

for $\hat{\mathbf{y}} = \texttt{vec}(\hat{\mathbf{Y}}^\mathsf{T})$, $\hat{\mathbf{x}} = \texttt{vec}(\hat{\mathbf{X}}^\mathsf{T})$, and $\hat{\mathbf{c}}'_* = \texttt{vec}(\hat{\mathbf{C}}_*^{'\mathsf{T}})$. Using the substitution

$$\mathbf{M} := \sum_{\hat{\psi}' \in \mathcal{P}'} \left( \mathbf{W}_{\hat{\psi}'}^\mathsf{T} \otimes \hat{\psi}'^\mathsf{T} - \mathbf{D}_{\hat{\psi}'} \otimes (\hat{\psi}'^\mathsf{T} \hat{\psi}'') \right)$$

and reforming the equation leads to the closed-form solution:

$$\hat{\mathbf{y}} = (\mathbf{I}_{nk} - \mathbf{M})^{-1} (\hat{\mathbf{x}} + \hat{\mathbf{c}}'_*)$$

which is defined if the spectral radius $\rho$ of $\mathbf{M}$ is smaller than 1. □

### One single directed edge type

Equation (23) simplifies further when we have just one single edge potential. In other words, we have a directed network and assume only one single type of edge whose meaning changes across the two directions (e.g., who follows some type of person on Twitter has a different meaning from who is followed by same type).

**Corollary 17** (One potential). *Let $k$ be the number of classes for each node in the graph, $\psi$ the $k \times k$-dimensional potential across an edge in the direction from source to target, $\hat{\mathbf{Y}}$ and $\hat{\mathbf{X}}$ the $n \times k$ dimensional final and explicit belief matrices, $\mathbf{W}$ the directed adjacency matrix, and $\mathbf{D}^{\text{in}}$ ($\mathbf{D}^{\text{out}}$) the in-degree (out-degree) matrices. Then the update equations simplify to:*

$$\hat{\mathbf{Y}} \leftarrow \hat{\mathbf{X}} + \hat{\mathbf{C}}'_* + \mathbf{W}^\mathsf{T} \hat{\mathbf{Y}} \hat{\psi}' + \mathbf{W} \hat{\mathbf{Y}} \hat{\psi}''^\mathsf{T} - \mathbf{D}^{\text{in}} \hat{\mathbf{Y}} \hat{\psi}''^\mathsf{T} \hat{\psi}' - \mathbf{D}^{\text{out}} \hat{\mathbf{Y}} \hat{\psi}' \hat{\psi}''^\mathsf{T} \quad (25)$$

*with*

$$\hat{\mathbf{C}}'_* := \frac{1}{k} \left( \mathbf{W}^\mathsf{T} \mathbf{1}_n \mathbf{1}_k^\mathsf{T} \hat{\psi}' + \mathbf{W} \mathbf{1}_n \mathbf{1}_k^\mathsf{T} \hat{\psi}''^\mathsf{T} \right)$$

**Corollary 18** (Convergence). *Update Eq. (25) converges if and only if $\rho(\mathbf{M}) < 1$ for*

$$\mathbf{M} = \mathbf{W}^\mathsf{T} \otimes \hat{\psi}'^\mathsf{T} + \mathbf{W} \otimes \hat{\psi}'' - \mathbf{D}^{\text{in}} \otimes (\hat{\psi}'^\mathsf{T} \hat{\psi}'') - \mathbf{D}^{\text{out}} \otimes (\hat{\psi}'' \hat{\psi}'^\mathsf{T})$$

### One symmetric, doubly stochastic potential

Recent work (Gatterbauer et al. 2015) derived a linearization of BP for the special case of one single symmetric, doubly stochastic edge potential that is used throughout the network (Recall that for such a potential all residual row and column sums are 0, and that by multiplying it by the number of classes it will be centered around 1). We can recover this special case from Corollary 15 and Proposition 16 with a slightly updated notation.

**Proposition 19** (One symmetric, doubly stochastic potential). *If the MRF contains only one single edge type with a symmetric doubly stochastic potential $\psi$, then the update equations simplify to:*

$$\hat{\mathbf{Y}} \leftarrow \hat{\mathbf{X}} + \mathbf{W} \hat{\mathbf{Y}} \hat{\psi}' - \mathbf{D} \hat{\mathbf{Y}} \hat{\psi}'^2$$

*At the same time, the closed form solution simplifies to:*

$$\hat{\mathbf{y}} = \left(\mathbf{I}_{nk} - \mathbf{W} \otimes \hat{\boldsymbol{\psi}}' + \mathbf{D} \otimes \hat{\boldsymbol{\psi}}'^2\right)^{-1} \hat{\mathbf{x}} \tag{26}$$

*Proof Proposition 19.* First, notice that for any symmetric potential $\boldsymbol{\psi} \in \mathbb{R}^{k \times k}$, $\hat{\boldsymbol{\psi}}' = \hat{\boldsymbol{\psi}}'' = \hat{\boldsymbol{\psi}}/k$, and hence $\mathbf{W}_{\hat{\boldsymbol{\psi}}'}^\intercal \hat{\mathbf{Y}} \hat{\boldsymbol{\psi}}' + \mathbf{W}_{\hat{\boldsymbol{\psi}}''}^\intercal \hat{\mathbf{Y}} \hat{\boldsymbol{\psi}}'' = (\mathbf{W}_{\hat{\boldsymbol{\psi}}'}^\intercal + \mathbf{W}_{\hat{\boldsymbol{\psi}}'})\hat{\mathbf{Y}} \hat{\boldsymbol{\psi}}'$. Thus, its adjacency matrix becomes symmetric. Since we only have one potential, we also have only one adjacency matrix $\mathbf{W}$. Furthermore, $\hat{\boldsymbol{\psi}}'^\intercal = \hat{\boldsymbol{\psi}}'$ and hence, $\hat{\boldsymbol{\psi}}''^\intercal \hat{\boldsymbol{\psi}}' = \hat{\boldsymbol{\psi}}'^2$.

Second, the constant term $\hat{\mathbf{C}}'_*$ disappears for doubly stochastic potentials. This follows from the proof of Lemma 12 and the fact that in any doubly stochastic matrix $\boldsymbol{\psi} \in \mathbb{R}^{k \times k}$, each column is centered around $\frac{1}{k}$.

This allows us to simplify Eq. (23) to

$$\hat{\mathbf{Y}} = \hat{\mathbf{X}} + \mathbf{W}\hat{\mathbf{Y}}\hat{\boldsymbol{\psi}}' - \mathbf{D}\hat{\mathbf{Y}}\hat{\boldsymbol{\psi}}'^2 \tag{27}$$

Similarly, applying above assumptions to our closed-form solution Eq. (24) leads to:

$$\hat{\mathbf{y}} = \left(\mathbf{I}_{nk} - \mathbf{W} \otimes \hat{\boldsymbol{\psi}}' + \mathbf{D} \otimes \hat{\boldsymbol{\psi}}'^2\right)^{-1} \hat{\mathbf{x}} \qquad \square$$

Notice that Eq. (27) and Eq. (26) are exactly the ones given by (Gatterbauer et al. 2015), except for slightly different notation. In particular, the authors choose to center the potential $\boldsymbol{\psi}$ around $1/k$, which is possible in the case that all nodes have the same number of classes $k$ (and thus all potentials are quadratic with $k \times k$ dimensions).[13] We also chose here to formulate Eq. (26) as $\hat{\mathbf{y}} = \texttt{vec}(\hat{\mathbf{Y}}^\intercal)$ instead of $\texttt{vec}(\hat{\mathbf{Y}})$ to keep the beliefs of same nodes adjacent in the resulting stacked vectors. Vectorizing instead the transpose, we get the exact original formulation:

$$\texttt{vec}(\hat{\mathbf{Y}}) = \left(\mathbf{I}_{nk} - \hat{\boldsymbol{\psi}}' \otimes \mathbf{W} + \hat{\boldsymbol{\psi}}'^2 \otimes \mathbf{D}\right)^{-1} \texttt{vec}(\hat{\mathbf{X}}) \tag{28}$$

Notice that in a slight abuse of notation, we used $\hat{\boldsymbol{\psi}}'$ in Theorem 4 for the sparse $k_{\text{tot}} \times k_{\text{tot}}$-square matrix, whereas we use it here for the single $k \times k$ recentered residual potential.

## D Illustrating examples

**Example 20** (Linearization). *Consider the network Fig. 7a consisting of nodes $N = \{1, 2, 3\}$. Node 1 has three classes, whereas nodes 2 and 3 have two classes. We have two edges, e.g., the edge between nodes 1 and 2 with a $3 \times 2$ potential $\boldsymbol{\psi}_{12}$. Fig. 7b illustrates Eq. (3). Notice that $\hat{\mathbf{c}}'_* = \hat{\boldsymbol{\psi}}'^\intercal \mathbf{k}$ with $\mathbf{k} = [\frac{1}{3}, \frac{1}{3}, \frac{1}{3}, \frac{1}{2}, \frac{1}{2}, \frac{1}{2}, \frac{1}{2}]^\intercal$ and $k_{\text{tot}} = 3 + 2 + 2$. Further notice that the matrix $\hat{\boldsymbol{\psi}}'^{\intercal 2}$ is block-diagonal (entries represent the echo modulation that a node receives through all its neighbors). In the following, we write $\langle \cdot \rangle_2$ for the projection of a stacked vector on the entries for node 2, e.g., $\langle \hat{\mathbf{y}} \rangle_2 = \hat{\mathbf{y}}_2$:*

$$\langle \hat{\mathbf{x}} \rangle_2 = \hat{\mathbf{x}}_2$$
$$\langle \hat{\mathbf{c}}'_* \rangle_2 = \tfrac{1}{3}\hat{\boldsymbol{\psi}}'^\intercal_{12}\mathbf{1}_3 + \tfrac{1}{2}\hat{\boldsymbol{\psi}}'^\intercal_{32}\mathbf{1}_2$$
$$\langle \hat{\boldsymbol{\psi}}'^\intercal \hat{\mathbf{y}} \rangle_2 = \hat{\boldsymbol{\psi}}'^\intercal_{12}\hat{\mathbf{y}}_1 + \hat{\boldsymbol{\psi}}'^\intercal_{32}\hat{\mathbf{y}}_3$$
$$\langle \hat{\boldsymbol{\psi}}'^{\intercal 2} \hat{\mathbf{y}} \rangle_2 = \underbrace{\left(\hat{\boldsymbol{\psi}}'^\intercal_{12}\hat{\boldsymbol{\psi}}'^\intercal_{21} + \hat{\boldsymbol{\psi}}'^\intercal_{32}\hat{\boldsymbol{\psi}}'^\intercal_{23}\right)}_{\hat{\boldsymbol{\psi}}_{2*}} \hat{\mathbf{y}}_2$$

*Then, the single update equation could also be written as several simultaneous update equations:*

$$\hat{\mathbf{y}}_1 \leftarrow \hat{\mathbf{x}}_1 + \langle \hat{\mathbf{c}}'_* \rangle_1 + \hat{\boldsymbol{\psi}}'^\intercal_{21}\hat{\mathbf{y}}_2 \qquad\qquad - \hat{\boldsymbol{\psi}}_{1*}\hat{\mathbf{y}}_1$$
$$\hat{\mathbf{y}}_2 \leftarrow \hat{\mathbf{x}}_2 + \langle \hat{\mathbf{c}}'_* \rangle_2 + \hat{\boldsymbol{\psi}}'^\intercal_{12}\hat{\mathbf{y}}_1 + \hat{\boldsymbol{\psi}}'^\intercal_{32}\hat{\mathbf{y}}_3 - \hat{\boldsymbol{\psi}}_{2*}\hat{\mathbf{y}}_2$$
$$\hat{\mathbf{y}}_3 \leftarrow \hat{\mathbf{x}}_3 + \langle \hat{\mathbf{c}}'_* \rangle_3 + \hat{\boldsymbol{\psi}}'^\intercal_{23}\hat{\mathbf{y}}_1 \qquad\qquad - \hat{\boldsymbol{\psi}}_{3*}\hat{\mathbf{y}}_3$$

**Example 21** (Repeating potentials). *We use Example 20 to also illustrate Proposition 14. Let $y_{sj}$ be the belief of node $s$ in class $j$. We create two belief matrices $\forall k \in K = \{2, 3\}$: $\hat{\mathbf{Y}}_2 = \begin{bmatrix} y_{21} & y_{22} \\ y_{31} & y_{32} \end{bmatrix}$, and $\hat{\mathbf{Y}}_3 = \begin{bmatrix} y_{11} & y_{12} & y_{13} \end{bmatrix}$. Thus, for example, the*

---

[13]"Row-recentering" and "column-recentering" as proposed in the present paper are more general forms of the centerings proposed in earlier work, which is necessary in order to deal with the general case of a non-quadratic and non-doubly stochastic potentials.

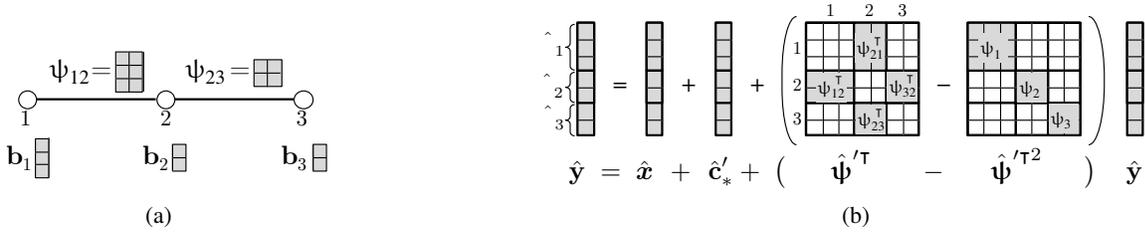

Figure 7: Example 20: (a): Network with 3 nodes and 2 edge potentials. (b): Resulting equation system where $\hat{\boldsymbol{x}}$ contains any prior beliefs and $\hat{\boldsymbol{c}}'_*$ is a vector that depends only on the graph and potentials.

> beliefs of node 2 are in row 1 of $\hat{\mathbf{Y}}_2$ (also written as $o(2) = 1$). We have four row-recentered matrices: $\hat{\boldsymbol{\psi}}'_{12}$, $\hat{\boldsymbol{\psi}}'_{21}$, $\hat{\boldsymbol{\psi}}'_{23}$, $\hat{\boldsymbol{\psi}}'_{32}$ with corresponding echo cancellation potentials (e.g., $\hat{\boldsymbol{\psi}}''^{\mathsf{T}}_{12}\hat{\boldsymbol{\psi}}'_{12}$ for $\hat{\boldsymbol{\psi}}'_{12}$), and appropriate adjacency and in-degree matrices. For example, $\hat{\boldsymbol{\psi}}'_{12} \in \mathbb{R}^{3\times 2}$ has $\mathbf{W}_{\hat{\boldsymbol{\psi}}'_{12}} = [\,1\ 0\,]$ where the first entry indicates an edge from node 1 to node 2. We illustrate next in detail:
>
> $$\mathbf{W}_{\hat{\boldsymbol{\psi}}'_{12}} = {}_1[\,\overset{2}{1}\ \overset{3}{0}\,],\ \mathbf{W}_{\hat{\boldsymbol{\psi}}'_{21}} = {}^2_3\!\begin{bmatrix}\overset{1}{1}\\0\end{bmatrix},\ \mathbf{W}_{\hat{\boldsymbol{\psi}}'_{23}} = {}^2_3\!\begin{bmatrix}\overset{2}{0}&\overset{3}{1}\\0&0\end{bmatrix},\ \mathbf{W}_{\hat{\boldsymbol{\psi}}'_{32}} = {}^2_3\!\begin{bmatrix}\overset{2}{0}&\overset{3}{0}\\1&0\end{bmatrix}$$
>
> $$\mathbf{D}^{\mathrm{in}}_{\hat{\boldsymbol{\psi}}'_{12}} = {}^2_3\!\begin{bmatrix}\overset{2}{1}&\overset{3}{0}\\0&0\end{bmatrix},\ \mathbf{D}^{\mathrm{in}}_{\hat{\boldsymbol{\psi}}'_{21}} = {}_1[\,\overset{1}{1}\,],\ \mathbf{D}^{\mathrm{in}}_{\hat{\boldsymbol{\psi}}'_{23}} = {}^2_3\!\begin{bmatrix}\overset{2}{0}&\overset{3}{0}\\0&1\end{bmatrix},\ \mathbf{D}^{\mathrm{in}}_{\hat{\boldsymbol{\psi}}'_{32}} = {}^2_3\!\begin{bmatrix}\overset{2}{1}&\overset{3}{0}\\0&0\end{bmatrix}$$
>
> We then get the following update equations:
>
> $$\hat{\mathbf{Y}}_2 \leftarrow \hat{\mathbf{X}}_2 + \hat{\mathbf{C}}'_{2*} + \begin{bmatrix}1\\0\end{bmatrix}\hat{\mathbf{Y}}_3\hat{\boldsymbol{\psi}}'_{12} - \begin{bmatrix}1&0\\0&0\end{bmatrix}\hat{\mathbf{Y}}_2\hat{\boldsymbol{\psi}}''^{\mathsf{T}}_{12}\hat{\boldsymbol{\psi}}'_{12}$$
> $$+ \begin{bmatrix}0&0\\1&0\end{bmatrix}\hat{\mathbf{Y}}_2\hat{\boldsymbol{\psi}}'_{23} - \begin{bmatrix}0&0\\0&1\end{bmatrix}\hat{\mathbf{Y}}_2\hat{\boldsymbol{\psi}}''^{\mathsf{T}}_{23}\hat{\boldsymbol{\psi}}'_{23}$$
> $$+ \begin{bmatrix}0&1\\0&0\end{bmatrix}\hat{\mathbf{Y}}_2\hat{\boldsymbol{\psi}}'_{32} - \begin{bmatrix}1&0\\0&0\end{bmatrix}\hat{\mathbf{Y}}_2\hat{\boldsymbol{\psi}}''^{\mathsf{T}}_{32}\hat{\boldsymbol{\psi}}'_{32}$$
> $$\hat{\mathbf{Y}}_3 \leftarrow \hat{\mathbf{X}}_3 + \hat{\mathbf{C}}'_{3*} + [\,1\ 0\,]\hat{\mathbf{Y}}_2\hat{\boldsymbol{\psi}}'_{21} - [\,1\,]\hat{\mathbf{Y}}_3\hat{\boldsymbol{\psi}}''^{\mathsf{T}}_{21}\hat{\boldsymbol{\psi}}'_{21}$$
>
> with
>
> $$\hat{\mathbf{C}}'_{2*} = \tfrac{1}{3}\begin{bmatrix}1\\0\end{bmatrix}[\,1\ 1\ 1\,]\hat{\boldsymbol{\psi}}'_{12} + \tfrac{1}{2}\begin{bmatrix}0&0\\1&0\end{bmatrix}\begin{bmatrix}1&1\\1&1\end{bmatrix}\hat{\boldsymbol{\psi}}'_{23} + \tfrac{1}{2}\begin{bmatrix}0&1\\0&0\end{bmatrix}\begin{bmatrix}1&1\\1&1\end{bmatrix}\hat{\boldsymbol{\psi}}'_{32}$$
> $$\hat{\mathbf{C}}'_{3*} = \tfrac{1}{2}[\,1\ 0\,]\begin{bmatrix}1&1\\1&1\end{bmatrix}\hat{\boldsymbol{\psi}}'_{21}$$

# E  Weighted edges in MRFs

The notion of "weights" on edges in MRFs is not defined and it is not immediately clear what an appropriate semantics would be. Here we give an a natural interpretation of edge weights in MRFs and derive a modification of the linearization to handle such weighted edges. We derive this interpretation starting from one single axiom:

**Axiom 22** (Edge weights). *An edge with weight $w \in \mathbb{N}$ behaves identically as $w$ parallel edges with the same potential but weight* 1.

From the original BP update equations Eq. (1) and Eq. (2), we see that two parallel edges carry the same messages, and that these two messages need to be multiplied to calculate the resulting messages and beliefs. It follows that these parallel edges behave identically to having one single edge with a new potential $\boldsymbol{\psi}_{st} \odot \boldsymbol{\psi}_{st}$, i.e., the result of element-wise multiplying the entries of the original potential. More generally, an edge with a potential $\boldsymbol{\psi}$ and weight $w$ is the same as an unweighted edge with a new potential $\boldsymbol{\psi}_w$ with entries entries $\psi_w(j,i) = \psi(j,i)^w$.

To see how weights affect our linearized formulation in terms of residuals, recall that $\psi(j,i) = 1 + \hat{\psi}(j,i)$. Therefore, $\psi(j,i)^w = \bigl(1 + \hat{\psi}(j,i)\bigr)^w = 1 + w\hat{\psi}(j,i) + \mathcal{O}(\hat{\psi}(j,i)^2)$. Under the assumption of small deviations from the center, we thus get: $\hat{\psi}_w(j,i) = w\hat{\psi}(j,i)$. Hence, *weights on edges simply multiply the residual potentials* in our linearized formulation. In other words, weights on an edges simply scale the coupling strengths between two neighbors.

It follows that Proposition 14 can be generalized to weighted networks by using weighted adjacency matrices $\mathbf{W}_{\hat{\boldsymbol{\psi}}'}$ with elements $A_{\hat{\boldsymbol{\psi}}'}(o(s), o(t)) = w > 0$ if the edge $s \to t$ with potential $\hat{\boldsymbol{\psi}}'$ and weight $w$ exists, and $W_{\hat{\boldsymbol{\psi}}'}(o(s), o(t)) = 0$ otherwise. In addition, each entry $D^{\text{in}}_{\hat{\psi}'}(o(t), o(t))$ in the block-diagonal matrix $\mathbf{D}^{\text{in}}_{\hat{\psi}'}$ is now the sum of the squared weights of edges to neighbors that are connected to $t$ via edge potential $\hat{\psi}$, instead of just the number of neighbors (recall that the echo cancellation goes back and forth, and notice again that the potentials work along the direction $s \to t$). After this modification, Proposition 14 can immediately be used for *weighted* graphs as well.

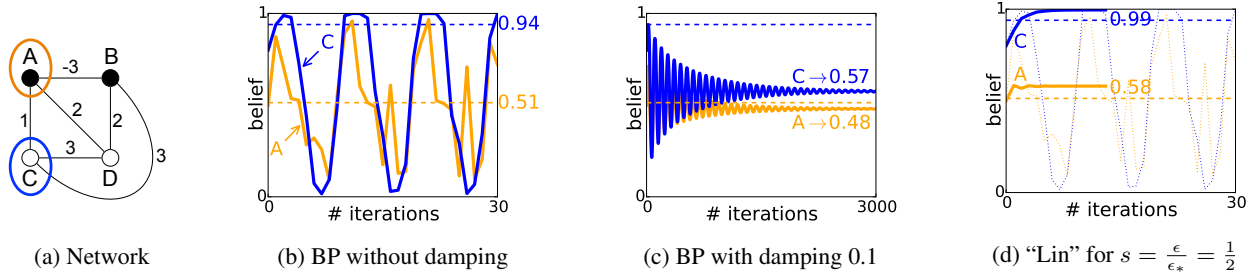

Figure 8: Example 24: Illustrating example for BP and its linearization "Lin". Details are given in the text.

**Example 23** (Edge weights). *We give here a small detailed example that shows the effects of weights for a potential whose entries are not really close to each other (i.e., the average entry is 1, however entries can diverge considerably from 1). We start with the potential $\boldsymbol{\psi} = \begin{bmatrix} 4 & 6 & 5 \\ 6 & 8 & 7 \end{bmatrix}$. By dividing all entries by 6, we get an equivalent potential that is centered around 1; and from this we get the residual and the row-recentered residual matrices:*

$$\boldsymbol{\psi} = \tfrac{1}{6}\begin{bmatrix} 4 & 6 & 5 \\ 6 & 8 & 7 \end{bmatrix}, \quad \hat{\boldsymbol{\psi}} = \tfrac{1}{6}\begin{bmatrix} -2 & 0 & -1 \\ 0 & 2 & 1 \end{bmatrix}, \quad \hat{\boldsymbol{\psi}}' = \tfrac{1}{18}\begin{bmatrix} 1 & -1 & 0 \\ 1 & -1 & 0 \end{bmatrix}$$

*The squared potential centered around 1 is then:* $\boldsymbol{\psi}_2 = \tfrac{3}{113}\begin{bmatrix} 4^2 & 6^2 & 5^2 \\ 6^2 & 8^2 & 7^2 \end{bmatrix}$. *And the residual and row-recentered residual matrices:*

$$\hat{\boldsymbol{\psi}}_2 \approx \begin{bmatrix} 0.575 & 0.044 & 0.336 \\ 0.044 & -0.699 & -0.300 \end{bmatrix}, \quad \hat{\boldsymbol{\psi}}'_2 \approx \begin{bmatrix} 0.085 & -0.091 & 0.006 \\ 0.121 & -0.127 & 0.006 \end{bmatrix}$$

*We can now compare the potential we get by multiplying the residual by 2, or by squaring the original potential and then recentering:*

$$2\hat{\boldsymbol{\psi}} \approx \begin{bmatrix} 0.111 & -0.111 & 0 \\ 0.111 & -0.111 & 0 \end{bmatrix}, \quad \hat{\boldsymbol{\psi}}'_2 \approx \begin{bmatrix} 0.085 & -0.091 & 0.006 \\ 0.121 & -0.127 & 0.006 \end{bmatrix}$$

*We see that the overall direction is correct, but there are considerable differences (e.g., $\approx 30\%$ relative difference for the first matrix entry: 0.111 vs. 0.085).*

*We next bring each entry in the potential closer to the center. Concretely, we reduce the deviation by one order of magnitude:*

$$\boldsymbol{\psi} = \tfrac{1}{6}\begin{bmatrix} 5.8 & 6.0 & 5.9 \\ 6.0 & 6.2 & 6.1 \end{bmatrix}, \quad \hat{\boldsymbol{\psi}} = \tfrac{1}{60}\begin{bmatrix} -2 & 0 & -1 \\ 0 & 2 & 1 \end{bmatrix}, \quad \hat{\boldsymbol{\psi}}' = \tfrac{1}{180}\begin{bmatrix} 1 & -1 & 0 \\ 1 & -1 & 0 \end{bmatrix}$$

*Now both versions are very close (e.g., $\approx 2\%$ relative difference for the first matrix entry: 0.0111 vs. 0.0109):*

$$2\hat{\boldsymbol{\psi}} \approx \begin{bmatrix} 0.0111 & -0.0111 & 0 \\ 0.0111 & -0.0111 & 0 \end{bmatrix}, \quad \hat{\boldsymbol{\psi}}'_2 \approx \begin{bmatrix} 0.0109 & -0.0110 & 0.00005 \\ 0.0113 & -0.0113 & 0.00005 \end{bmatrix}$$

## F   Illustrating examples

**Example 24** (Convergence). *We illustrate the different convergence behaviors of BP and our formalism in a graph with several different potentials. The scenario is a variant of an example given by (Heskes 2002) of a 4-clique graph with weights on the edges (see Fig. 8a). The weights are used to entry-wise exponentiate the entries of the potential $\boldsymbol{\psi} = \begin{bmatrix} 4 & 1 \\ 1 & 4 \end{bmatrix}$ before normalization: $\psi_{(w)}(i,j) = \bigl(\psi(i,j)\bigr)^w$. For example, a weight 2 leads to $\boldsymbol{\psi}_{(2)} = \begin{bmatrix} 16 & 1 \\ 1 & 16 \end{bmatrix}$, and a weight $-3$ leads to $\boldsymbol{\psi}_{(-3)} = \begin{bmatrix} \tfrac{1}{64} & 1 \\ 1 & \tfrac{1}{64} \end{bmatrix}$. The graph has 2 nodes A and B with prior beliefs: $\boldsymbol{x}_A = \boldsymbol{x}_C = [0.8, 0.2]^\mathsf{T}$.*

*Figure 8b shows the beliefs for nodes A and C for various iterations of BP. The dashed lines show the maximum marginal (MM) distribution as determined by complete iteration. We see that BP has a somewhat erratic cyclic behavior and does not converge. Figure 8c shows a variant that uses damping (Koller and Friedman 2009, ch. 11.1), a method that is often used to make BP converge when it otherwise would not. Damping with 0.1 (meaning an updated message is a linear combination of 90% the prior message and 10% of newly calculated values) is able to dampen the the behavior, but convergence happens only after 1,000s of iterations, and even then the maximum marginal for node A is wrong (0.48 leads to choosing class 2, vs. 0.51 leads to choosing class 1). Furthermore, after replacing $\boldsymbol{\psi}$ in our example with $\begin{bmatrix} 8 & 1 \\ 1 & 8 \end{bmatrix}$, BP will not converge anymore for any damping factor and the fixed points of BP are unstable.*

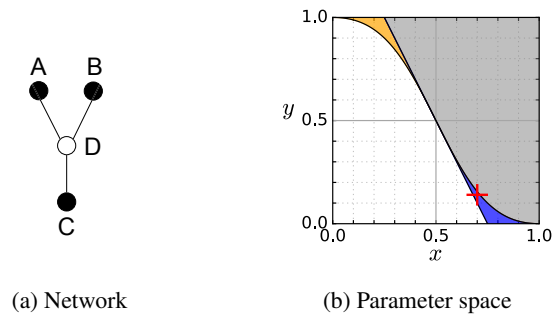

(a) Network     (b) Parameter space

Figure 9: Example 25: (a) Example network. (b) BP and the linearization lead to different predictions for the colored parameter choices.

*Figure 8d illustrates the convergence of the linearization: The convergence boundary is $\epsilon_* = 0.3109$. By choosing a convergence parameter $s = \frac{\epsilon}{\epsilon_*} = \frac{1}{2}$, the values converge after a few iterations. In addition, the MM beliefs coincide with the exact solution.*

**Example 25** (Errors of linearization). *We know that the approximation becomes worse if values are not close to the centering point. We give here a simple example to illustrate the impact this happens on the actual node classification. Consider the network in Fig. 9a with $n = 4$ nodes and each node having $k = 2$ classes. Let $\boldsymbol{x}_A = \boldsymbol{x}_B = [x, \bar{x}]^\mathsf{T}$ and $\boldsymbol{x}_C = [y, \bar{y}]^\mathsf{T}$, where $\bar{x} \coloneqq 1 - x$. We will calculate $\mathbf{y}_D = [z, \bar{z}]^\mathsf{T}$ as a function of $x$ and $y$ (representing the prior beliefs at the nodes in the respective classification) and then assign the class 1 to D if $z > \bar{z}$, or equally $z > 0.5$.*
*For BP, the following condition needs to hold for D to be labeled as class 1:*

$$x^2 y > \bar{x}^2 \bar{y}$$

*In contrast, for the linearization, the following condition needs to hold (notice that we formulated the condition on the residuals):*

$$2\hat{x} + \hat{y} > 0$$

*Both conditions are equivalent close to the centering point $0.5$, however diverge further away from that point. Figure 9b illustrates 4 regimes for parameters $x$ and $y$: In the large gray area (upper right) both algorithms are give identical predications and assign class 1 (analogously for the white area in the lower left). But in the blue area (lower right), BP assigns class 2 whereas its linearization 1 (analogously for the yellow area in the upper left). Now consider the red cross marking the point $(x = 0.7, y = 0.14)$ in the blue regime. BP calculates $z = 0.47$ whereas the approximation calculates $z = 0.54$.*

## G  Existing graph generators and hardness of node labeling

There is a large body of work that proposes various realistic synthetic generative graph models. However, almost all of this work assumes unlabeled graphs. While one could use these existing graph generators to have realistic graphs, one cannot easily take a graph and then label the nodes according to some desired compatibility matrix. In fact, this problem is NP-hard.

**Proposition 26** (Labeling with potentials). *Given a graph $G(V, E)$. Finding labels $\ell : v \in V \to [k]$ so that the labels follow a given stochastic affinity matrix $\boldsymbol{\psi}$ (where $\psi(i, j)$ determines the average fraction of nodes of class $j$ connected to a node of class $i$) is NP-hard.*

*Proof Proposition 26.* Membership in NP follows from the fact that we can easily verify a solution by calculating the average neighbor-to-neighbor relations in a labeled graph. We use a simple reduction from the problem of Graph 3-colorability. Graph 3-colorability is the question of whether there exists a labeling function $k : v \to \{1, 2, 3\}$ such that $k(u) \neq k(v)$ whenever $(u, v) \in E$ for a given graph $G = (V, E)$ and is well known to be NP-hard (Stockmeyer 1973). Assume now that we have a method that allows us to label any graph $G(V, E)$ following the heterophily matrix $\boldsymbol{\psi} = \frac{1}{2}\begin{bmatrix} 0 & 1 & 1 \\ 1 & 0 & 1 \\ 1 & 1 & 0 \end{bmatrix}$, i.e., neighboring nodes never have the same label. It follows immediately that such a solution would also be solution to graph 3-colorability. □

We thus need graph generators which generate both the graph topology (i.e., $\mathbf{W}$) and the node labels (i.e., $\mathbf{X}$) in the same process. We know of only two papers that have proposed graph generators that generate labeled data in the process: the early work by (Sen and Getoor 2007) and the very recent work by (Lee et al. 2015). Neither graph generator is available. In addition, neither of the papers gives a way to know the "ground truth" actual potential matrix that was used to label data (e.g., (Lee et al. 2015) suggests this as future work).

We therefore had to implement our own synthetic graph generator with the additional design decision that any potential matrix can be "planted" as *exact graph property*. This allows us to separate the concern between (1) how does our method work on graphs with certain properties, (2) what is the variation in properties of a given generative model. By planting exact properties (instead of expected properties) we can focus on question (1) only. The random graph generator is described in detail in (Gatterbauer 2016).